\documentclass{article}

\PassOptionsToPackage{numbers, compress}{natbib}
\usepackage[preprint]{neurips_2026}

\usepackage[utf8]{inputenc} % allow utf-8 input
\usepackage[T1]{fontenc}    % use 8-bit T1 fonts
\usepackage{hyperref}       % hyperlinks
\usepackage{url}            % simple URL typesetting
\usepackage{booktabs}       % professional-quality tables
\usepackage{amsfonts}       % blackboard math symbols
\usepackage{nicefrac}       % compact symbols for 1/2, etc.
\usepackage{microtype}      % microtypography
\usepackage{xcolor}         % colors
\usepackage{graphicx}
\usepackage{amsmath} 
\usepackage{bm}
\usepackage{enumitem}
\usepackage{multirow}
\usepackage{subcaption}
\usepackage{xcolor}
\usepackage{colortbl}
\usepackage{wrapfig}
\usepackage{array}
\usepackage{makecell}

\newcommand{\std}[1]{\textsubscript{\scriptsize$\pm$#1}}

\title{Structural Self-Teaching for Compositional Generation in Unified Multimodal Models}

\author{
Jaeyoo Park \textsuperscript{\normalfont 1}~~~~~~~~Bohyung Han\textsuperscript{\normalfont 1,2} \\\\ 
   \textsuperscript{1}Seoul National University \\
   \textsuperscript{2}Samsung Electronics  \\ 
   \texttt{\{bellos1203,\,bhhan\}@snu.ac.kr}
}

\begin{document}

\maketitle

%-------------------------------------------------------------------
%		Abstract
%-------------------------------------------------------------------
% !TEX root = ../main.tex

\begin{abstract}
We propose Structural Self-Teaching (SST), a novel framework that enhances compositional image generation in diffusion-based unified multimodal models (UMMs) by leveraging their internal understanding path as a source of structural supervision. 
The generation path of UMMs often struggles with dense compositional prompts, leading to missing objects, attribute binding failures, and spatial errors.
In contrast, their understanding path localizes the same phrases in images that satisfy these prompts.
SST converts this phrase-level grounding into training-time structural signals for the generation path.
Specifically, we extract phrase-conditioned support masks from internal attention maps derived from the understanding path and employ them through two mechanisms.
Phrase-local contrastive alignment uses these masks as pooling regions and aligns each generated phrase region with the same region of the target image.
A structural scaffold builds a phrase-level layout from the same masks and modulates an intermediate spatial state of the generation path during training. 
The scaffold is phased out during training and removed at inference.
SST requires no external structural annotations for training and no additional input at test time.
Extensive experiments across multiple UMMs demonstrate that SST improves compositional fidelity over LoRA fine-tuning on the same data, with no inference-time cost.
Project page is available at \url{https://bellos1203.github.io/sst}
\end{abstract}

%%%%%%%%% BODY TEXT
%-------------------------------------------------------------------
%		Introduction
%-------------------------------------------------------------------
% !TEX root = ../main.tex

\section{Introduction}
\label{sec:introduction}

% 1) Unified Multimodal Models & Remaining challenge
Unified multimodal models (UMMs) integrate image understanding and generation within a single architecture, and recent systems have shown strong perception performance together with competitive generation quality~\cite{deng2025emerging,wu2025janus,chen2025janus,xie2024show,xie2025show,chen2025blip3}.
Despite recent progress, UMMs still struggle to generate images that faithfully satisfy dense compositional prompts.
Given prompts that specify multiple objects, attributes, and relations, state-of-the-art UMMs often miscount instances, swap attributes across entities, and drop referenced phrases entirely.
These failures intensify as prompts become compositionally denser, motivating training methods that explicitly account for phrase-level structure.

\begin{figure}[t!]
	\centering
	\includegraphics[width=\textwidth]{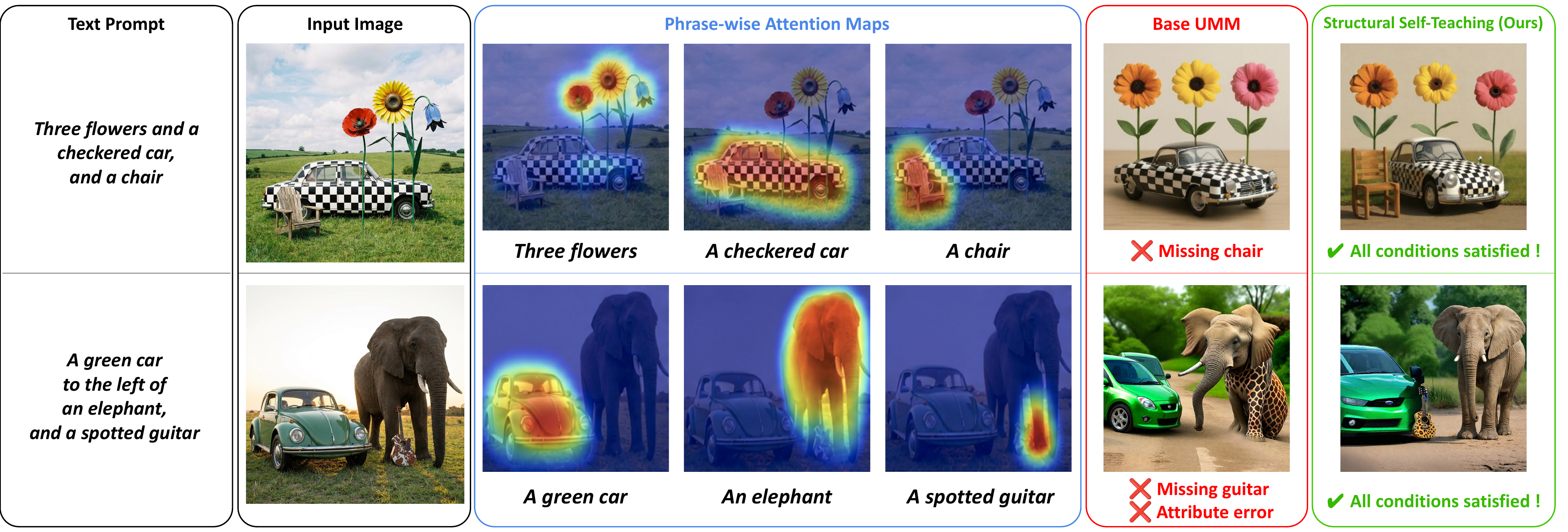}
	%\vspace{-1mm}
	\caption{
	Unified multimodal models can localize phrases they fail to generate.
        Given an image-text pair that satisfies a compositional prompt, the understanding path produces phrase-wise attention maps that localize the requested concepts.
        However, generation from the same prompt can omit objects or entangle attributes, as shown by the base UMM.
        Structural self-teaching transfers these internal supports during training, improving phrase coverage and compositional fidelity.
	}
	\vspace{-0.3cm}
\label{fig:teaser}
\end{figure}

% 2) Motivation : understanding path retains phrase-level structure that we can use.
UMMs also provide an internal source of phrase-level structure through their understanding path.
Given a training image and its caption, attention between caption and image tokens in the understanding path often localizes individual noun phrases, such as ``three flowers,'' ``a checkered car,'' or ``a chair,'' in spatially coherent image regions (Figure~\ref{fig:teaser}), consistent with prior observations in vision-language models~\cite{park2023multi, kang2025your}. 
These phrase-conditioned attention maps provide soft spatial supports over visual tokens. 
Section~\ref{sec:motivation} evaluates whether such supports remain usable under increasing compositional density. 
We use this internal grounding as a source of structural supervision during generation training, without relying on external annotations.

% 3) Our approach : structural self-teaching
To expose this internal structure to the generation path, we propose \emph{structural self-teaching} for diffusion-based UMMs, a framework that turns the understanding path into an internal structural teacher for its generation path. 
The central idea is to extract phrase-level grounding from the understanding path and convert it into structural signals for training the generation pathway. 
All structural cues are derived from the model itself, requiring neither external structural annotations nor inference-time intervention.

% 4) Two specific mechanisms
Phrase-level structure reaches the generation path through two routes.
The first route is phrase-local contrastive alignment.
It uses phrase-conditioned support masks as pooling regions for representation learning.
It pulls each phrase-specific region in the generated representation toward the corresponding region in the target representation, while pushing it away from other phrase-specific regions in the same prompt. 
The second is a structural scaffold, which converts the same support masks into a dense phrase-level canvas and modulates intermediate spatial states during training. 
Scaffold dropout prevents reliance on the canvas. The scaffold branch is removed at inference.
Together, the two mechanisms transfer understanding-side structure through both phrase-level representation alignment and spatial modulation during denoising.

% Contributions
The main contributions of our paper are summarized as follows:
%\vspace{-2mm}
%
\begin{itemize}[label=$\bullet$]
    \item We propose \emph{structural self-teaching}, a training framework that extracts phrase-conditioned support masks from the understanding path of a UMM and reuses them to supervise its generation path without external structural annotations or test-time inputs.
    \item We introduce phrase-local contrastive alignment, which leverages the extracted masks to align phrase-specific generated and reference visual representations while preventing semantic collapse among distinct phrase-level features within the same prompt.
    \item We develop an attenuated structural scaffold that converts these supports into a dense phrase-level canvas for training-time modulation, with a keep probability that decays over training.
    \item The proposed approach demonstrates consistent improvements in compositional fidelity across multiple UMMs, effectively addressing complex object, attribute, and spatial constraints without modifying the inference-time interface.
\end{itemize}

The rest of the paper is organized as follows.
Section~\ref{sec:related_works} reviews the related work, followed by our motivation in Section~\ref{sec:motivation}.
The details of our approach are described in Section~\ref{sec:method}, and the experimental results are presented in Section~\ref{sec:experiments}.
Finally, we conclude this paper in Section~\ref{sec:conclusion}.

%-------------------------------------------------------------------
%		Related Work
%-------------------------------------------------------------------
% !TEX root = ../main.tex

\section{Related Works}
\label{sec:related_works}

\subsection{Unified Multimodal Models}
\label{sec:UMM}
% UMMs
Unified multimodal models (UMMs) integrate visual understanding and generation within a single backbone, and generate images either autoregressively over discrete tokens~\cite{wu2025janus,chen2025janus,xie2024show,team2024chameleon} or through a diffusion process~\cite{deng2025emerging,zhou2024transfusion,xie2025show,chen2025blip3}. 
Despite competitive performance on both pathways, UMMs exhibit a persistent asymmetry in which they understand more reliably than they generate, prompting a body of post-training work aimed at closing this gap. 
HermesFlow~\cite{yang2025hermesflow} first quantifies this asymmetry and proposes preference optimization (Pair-DPO) using self-curated preference pairs from both pathways, aligning the two capabilities at the prompt level.
SRUM~\cite{jin2026srum} uses the model’s own understanding module to derive global and local rewards for reward-weighted generation training.
RecA~\cite{xie2025reconstruction} instead conditions the UMM on its own understanding embeddings and reconstructs the input image, providing dense pixel-level supervision without captions.
Our method also leverages internal signals from the understanding path, but targets compositional structure rather than reward-based post-training or pixel-level reconstruction: we transfer phrase-conditioned spatial supports through a phrase-local contrastive objective, directly addressing the entanglement and omission failures that emerge as compositional complexity grows.

% \vspace{-0.2cm}
\subsection{Visual Grounding and Attention in MLLMs}
\label{sec:attn_mllms}
Multimodal transformers encode useful structural signals in their attention patterns~\cite{li2021align, park2023multi, park2024hierarchical, kang2025your}. 
In multimodal encoders, word-conditional attention localizes caption words well enough to guide representation learning without region annotations~\cite{park2023multi}.
In frozen LVLMs, Kang et al.~\cite{kang2025your} show that a small subset of attention heads acts as localization heads and yield strong visual grounding without grounding-specific training, indicating that phrase-conditioned structure already emerges inside the model. 
At the same time, Visual Attention Sink~\cite{kang2025see} observes that a few uninformative tokens absorb disproportionate attention mass, meaning raw maps mix signal with systematic artifacts. 
The practical question is therefore how to extract this structure selectively. 
We adopt head scoring and sink suppression from previous works~\cite{kang2025your, kang2025see}, and extend them by scoring and selecting heads per phrase rather than globally—since different phrases in the same image are grounded by different heads—and by replacing hard thresholding with soft, temperature-controlled aggregation so that informative sub-threshold maps still contribute to a dense training target.

% \vspace{-0.2cm}
\subsection{Compositional Text-to-Image Generation}
\label{sec:comp_t2i}
Even strong generators struggle with compositional prompts, exhibiting missing entities, incorrect attribute binding, unreliable counting, and broken spatial relations—failures that recent benchmarks such as TIFA~\cite{hu2023tifa}, GenEval~\cite{ghosh2023geneval}, and GenEval2~\cite{kamath2025geneval} make quantitatively visible. 
Existing remedies act either during sampling or through added inputs.
Prompt-to-Prompt~\cite{hertz2022prompt} and Attend-and-Excite~\cite{chefer2023attend} manipulate cross-attention in each sampling run and leave the trained model unchanged.
ControlNet~\cite{zhang2023adding} and GLIGEN~\cite{li2023gligen} condition generation on external structure such as edges, depth, or bounding boxes, and LayoutGPT~\cite{feng2023layoutgpt} produces such layouts with an LLM planner.
These methods require a structural input beyond the prompt, from the user or from a separate planner.
Our approach instead regularizes the generation path during training with structure from the model’s own understanding path. 
It adds no inference-time step and no input beyond the prompt.

%-------------------------------------------------------------------
%		Preliminary Analysis
%-------------------------------------------------------------------
% !TEX root = ../main.tex

%\iffalse
\begin{figure}[t]
    \centering
    \begin{subfigure}{0.24\textwidth}
        \centering
        \includegraphics[width=\linewidth]{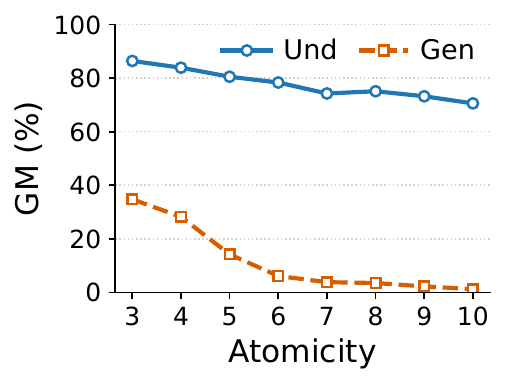}
        \par%\vspace{0.3em}
        \includegraphics[width=\linewidth]{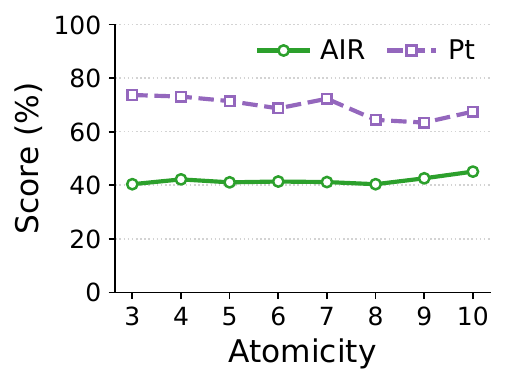}
        \caption{BLIP-3o 4B}
    \end{subfigure}
    \hfill
    \begin{subfigure}{0.24\textwidth}
        \centering
        \includegraphics[width=\linewidth]{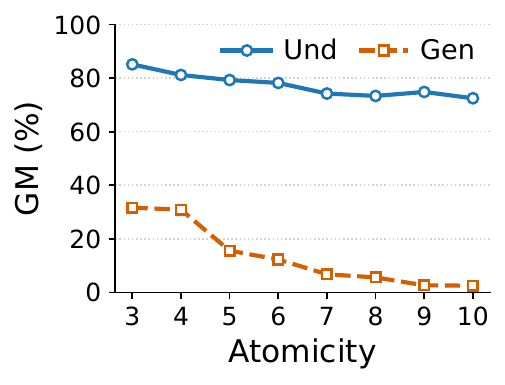}
        \par%\vspace{0.3em}
        \includegraphics[width=\linewidth]{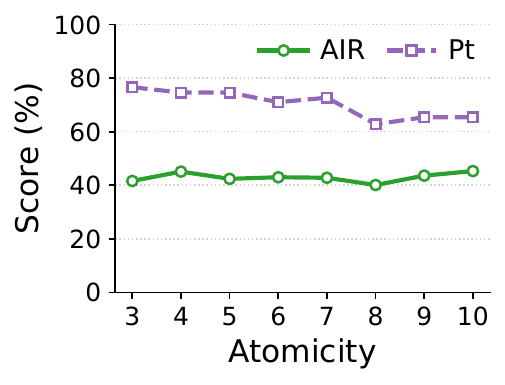}
        \caption{BLIP-3o 8B}
    \end{subfigure}
    \hfill
    \begin{subfigure}{0.24\textwidth}
        \centering
        \includegraphics[width=\linewidth]{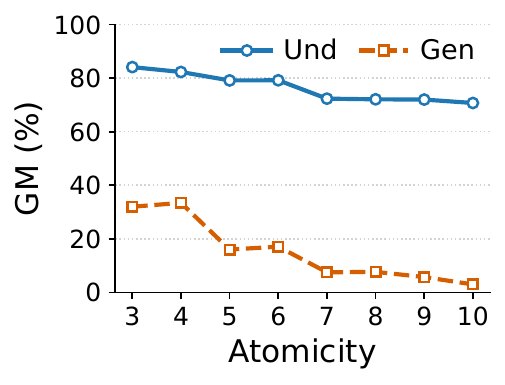}
        \par%\vspace{0.3em}
        \includegraphics[width=\linewidth]{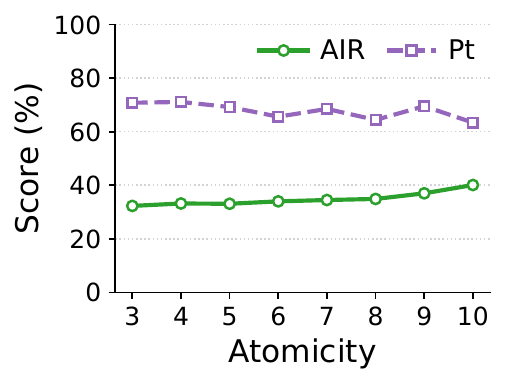}
        \caption{Show-o2 7B}
    \end{subfigure}
    \hfill
    \begin{subfigure}{0.24\textwidth}
        \centering
        \includegraphics[width=\linewidth]{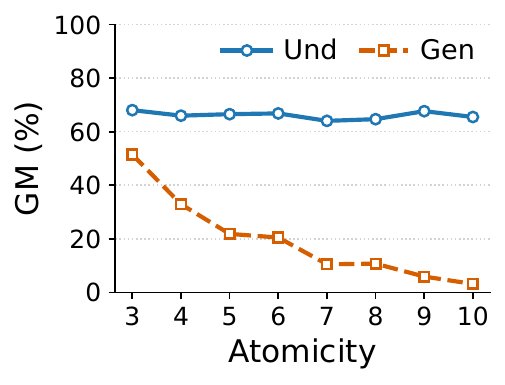}
        \par%\vspace{0.3em}
        \includegraphics[width=\linewidth]{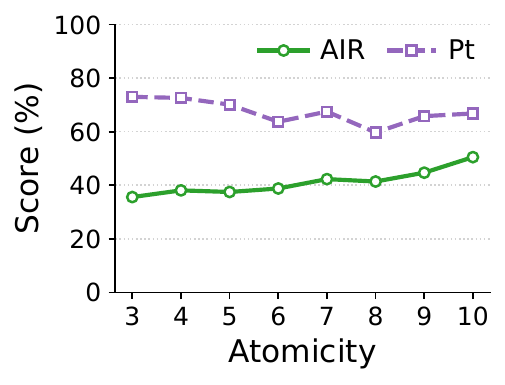}
        \caption{BAGEL}
    \end{subfigure}
\vspace{-1mm}
    \caption{
    Diagnostic of the understanding-path supervision source across four UMM backbones. 
    \textbf{Top row.} Prompt-only generation scores (Gen) decrease sharply with atomicity, while understanding-path scores on oracle images (Und) remain comparatively stable. 
    \textbf{Bottom row.} Activation-inside ratio (AIR) and center-of-mass pointing accuracy (Pt) of extracted phrase supports against Grounded-SAM pseudo-references remain within narrow bands across atomicity. 
    The top row is not a direct comparison between Und and Gen, see Section~\ref{sec:motivation}.
    }
    \vspace{-4mm}
    \label{fig:motivation}
\end{figure}
%\fi

\section{Motivation: Using Understanding Path as Structure Supervision}
\label{sec:motivation}

%Goal of analysis
Our framework uses the understanding path of a UMM as an internal source of phrase-level structural supervision. 
We therefore first examine whether this path provides usable compositional cues under dense prompts, considering two axes. 
The first is whether prompt-specified visual facts remain measurable when the corresponding image evidence is available, and the second is whether these facts are spatially grounded enough to define phrase-level supports.

% Setup for analysis
We use the atomic questions from GenEval2~\cite{kamath2025geneval}, which decompose each prompt into object, attribute, count, and spatial facts. 
Since GenEval2 provides no oracle images, we curate them through model-based generation and editing to satisfy all atomic conditions (see Appendix~\ref{app:oracle} for details).
These oracle images serve only as diagnostic inputs for probing the understanding path, and are not used for training or as a generation benchmark. 
Given each oracle image and prompt, the understanding path answers the corresponding atomic questions, and we separately report prompt-only generation scores under the GenEval2 protocol to indicate the compositional difficulty of the same prompts. 
We emphasize that Figure~\ref{fig:motivation} serves as a diagnostic of the source signal used by our method, not as a direct comparison between perception and generation.

% Quantitative evidence from Diagnostic results
Across BLIP-3o~\cite{chen2025blip3}, Show-o2~\cite{xie2025show}, and BAGEL~\cite{deng2025emerging}, prompt-only generation scores decrease sharply as atomicity increases, dropping to near-zero on the densest prompts (Figure~\ref{fig:motivation}, top row). 
In contrast, understanding-path scores on the corresponding oracle images remain comparatively stable across the same atomicity range, with only mild degradation at high atomicity. 
This stability holds across all evaluated backbones, despite their differences in architecture and scale.
The understanding path therefore retains measurable compositional information even under high atomicity, supporting its use as an internal source of structural supervision.

% Spatial grounding
We next examine the second axis, whether the source signal is spatially grounded and stable across compositional density. 
We extract phrase-conditioned support masks from the understanding path (following the procedure detailed in Section~\ref{sec:method-extraction}) and compare them against pseudo-reference masks obtained by running Grounded-SAM~\cite{ren2024grounded} on the same noun phrases. 
Across atomicity levels 3–10, the activation-inside ratio and center-of-mass pointing accuracy remain within narrow bands across all four backbones, with no systematic degradation as compositional density grows (Figure~\ref{fig:motivation}, bottom row). 
Since Grounded-SAM pseudo-references are themselves imperfect, we focus on stability under increasing atomicity rather than on absolute localization accuracy. 
These supports form spatially coherent regions over visual tokens, which we use as pooling regions for phrase-local alignment and as spatial carriers for a structural scaffold during generation training.

%-------------------------------------------------------------------
%		Method
%-------------------------------------------------------------------
% !TEX root = ../main.tex

\section{Structural Self-Teaching for Compositional Generation}
\label{sec:method}

%\iffalse
\begin{figure}[t!]
	\centering
	\includegraphics[width=0.98\textwidth]{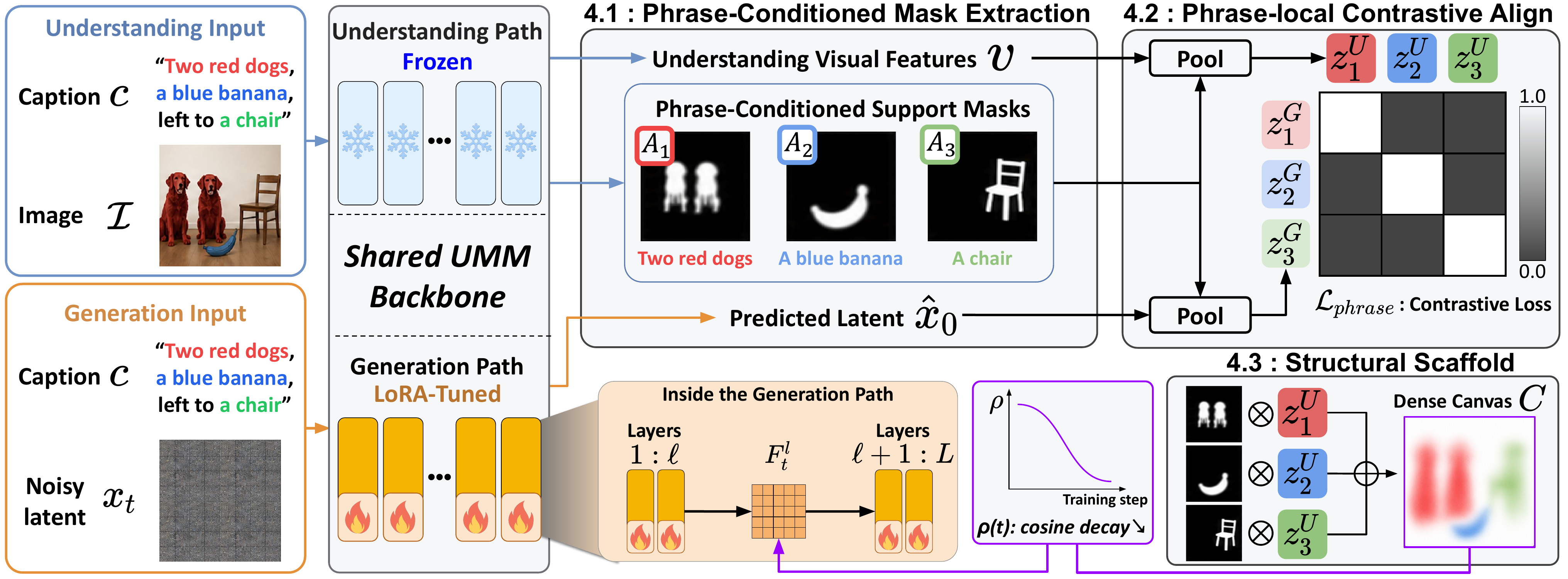}
	\caption{
	Overview of structural self-teaching.
	Phrase-conditioned support masks \(\{A_p\}\) extracted from the frozen understanding path provide two training-time signals for the LoRA-tuned generation path: phrase-local contrastive alignment between reference and generated phrase representations, and a structural scaffold that modulates an intermediate spatial feature \(F_t^\ell\). 
    The scaffold is applied with sample-wise cosine-decayed dropout and removed at inference.
	}
	%\vspace{-3mm}
\label{fig:framework}
\end{figure}
%\fi

% 4.0 Overview
%\subsection{Overview}
%\label{sec:overview}
The proposed Structural Self-Teaching (SST) framework transfers phrase-level structural cues from the understanding path of a UMM to its generation path during training (Figure~\ref{fig:framework}). 
Given an image and its corresponding caption, we first extract noun phrases and derive phrase-conditioned support masks from internal text-to-image attention maps within the understanding path (Section~\ref{sec:method-extraction}). 
The support masks reach the generation path through two routes. 
Phrase-local contrastive alignment uses them as pooling regions for a representation-level target (Section~\ref{sec:method-contrastive}).
The structural scaffold uses them as a spatial canvas that modulates an intermediate generation state during training (Section~\ref{sec:method-scaffold}).
Attenuated scaffold dropout phases the scaffold out during training. 
The scaffold branch is removed at inference, so the original generation interface requires no additional input.
The final objective combines the model's native denoising loss with the phrase-local contrastive loss (Section~\ref{sec:method-objective}).

% 4.1 Phrase-Conditioned Mask Extraction from Understanding Path
\subsection{Phrase-Conditioned Support Mask Extraction} % from the Understanding Path} 
\label{sec:method-extraction}

% Overview of extraction
To construct a reliable teacher signal from the understanding path, we extract a soft spatial support mask for each noun phrase in the caption. 
Our extraction procedure builds upon the attention-head scoring criteria of \cite{kang2025your}, originally proposed for zero-shot visual grounding, but we introduce two critical adaptations for compositional generation. 
First, we perform candidate scoring and selection per phrase rather than identifying single head set for the entire prompt; this reflects the observation that different phrases within the same complex scene are often grounded by distinct sets of heads. 
Second, we replace rigid threshold-based selection with a soft, temperature-controlled aggregation, ensuring that informative sub-threshold maps still contribute to the final spatial support.

% Specific method
Given an image $\mathcal{I}$ and its corresponding caption $c$, we extract noun phrases $\{p_n\}_{n=1}^{N}$ from $c$ and collect text-to-image attention from every layer-head pair of the understanding path. 
This produces a candidate pool $A_{p,\ell,h} \in \mathbb{R}^{M}$ for each phrase $p$ and layer-head pair $(\ell, h)$, where $M$ denotes the number of image tokens. 
Following~\cite{kang2025see}, we preprocess each candidate map to suppress background activations and visual attention sinks, then rank these resulting candidate maps within each phrase based on a combined score $s_{p, \ell, h}$ of relevance and spatial plausibility as:
\begin{equation}
    s_{p, \ell, h} = s^{\mathrm{sum}}_{p, \ell, h} + s^{\mathrm{ent}}_{p, \ell, h},
\end{equation}
where the relevance term $s^{\mathrm{sum}}_{p, \ell, h}$ and the spatial plausibility term $s^{\mathrm{ent}}_{p, \ell, h}$ are both normalized within the phrase-specific candidate pool to ensure comparability.

Let $\bar{A}_{p, k}$ be the $k$-th ranked candidate map after the aforementioned normalization. 
The final phrase-conditioned support mask $A_p$ is then constructed by softly aggregating the top-$K$ candidate maps with temperature-controlled weights:
\begin{equation}
    A_{p} = \text{Norm}\left( \sum_{k=1}^{K} w_{p, k} \, \bar{A}_{p, k} \right),
    \qquad
    w_{p, k} = \frac{\exp(s_{p, k} / \tau)}{\sum_{k'=1}^{K} \exp(s_{p, k'} / \tau)},
\end{equation}
where the temperature $\tau > 0$ controls the sharpness of the aggregation, and $\text{Norm}(\cdot)$ denotes a final sum-to-one renormalization step to ensure the mask values stay within a valid range for the subsequent alignment stage.

The resulting $A_p$ serves as a phrase-conditioned soft spatial support, rather than a binary segmentation or instance mask. 
By reliably providing coarse localization and separating spatial responsibilities across phrases, $A_p$ acts as a frozen, structurally-grounded teacher that guides the generation path.
Implementation details and exact forms of $s^{\mathrm{sum}}$ and $s^{\mathrm{ent}}$ are provided in Appendix~\ref{app:mask-extraction}.
Example visualizations are shown in Appendix~\ref{app:mask_viz}.

% 4.2 Phrase-Local Contrastive Loss
\subsection{Phrase-local Contrastive Alignment}
\label{sec:method-contrastive}

We leverage the phrase-conditioned support masks $A_p$ as pooling operators to provide representation-level supervision. 
Given a caption $c$, its corresponding ground-truth image $I$, and a sampled timestep $t$, the generation path predicts a denoised latent $\hat{x}_0$ from $x_t$ conditioned on $c$.
We encode $I$ into reference visual features $v \in \mathbb{R}^{M \times D}$. % (Appendix~\ref{app:backbones}). 
For each noun phrase $p$ in the caption, we derive a reference visual representation $z_p^U$ from the target image and a generated visual representation $z_p^G$ from the denoised latent via mask-weighted pooling:
\begin{equation}
    z_p^U = \mathrm{Pool}(v, A_p), \qquad
    z_p^G = g_\phi(\mathrm{Pool}(\hat{x}_0, A_p)),
    \label{eq:contrastive}
\end{equation}
where $\mathrm{Pool}(\cdot, A_p)$ denotes mask-weighted spatial averaging, with $A_p$ resized to match the spatial resolution of the respective feature maps. 
The projection head $g_\phi$ maps the pooled features from the generation path into the visual feature space of $v$.
The choice of features and projection is backbone-specific (Appendix~\ref{app:backbones}).

The alignment objective pulls each generated representation toward its own reference and away from the references of other phrases in the same prompt:
\begin{equation}
    \mathcal{L}_{\mathrm{phrase}} = -\frac{1}{N} \sum_{p=1}^{N} \log \frac{ \exp(\mathrm{sim}(z_p^G, z_p^U) / \tau_{\text{align}}) }{ \sum_{q=1}^{N} \exp(\mathrm{sim}(z_p^G, z_q^U) / \tau_{\text{align}})},
    \label{eq:contrastive_obj}
\end{equation}
where $\mathrm{sim}(\cdot, \cdot)$ denotes cosine similarity and $\tau_{\text{align}} > 0$ is a temperature. 
The positive term matches the content generated in each phrase region to the corresponding region of the reference image.
The in-prompt negatives keep distinct phrases from collapsing onto one representation, which targets the attribute leakage and object fusion observed in dense prompts.

% 4.3 Construction of Structural Scaffold
\subsection{Attenuated Structural Scaffold}
\label{sec:method-scaffold}

The structural scaffold modulates an intermediate spatial state of the generation path with a phrase-level layout during training.
The layout assigns each phrase’s reference representation to its support region. 
Phrase-local contrastive alignment specifies what each phrase region should contain, but it reaches the generation path only through the gradient of a pooled loss. 
We attenuate the scaffold over training and remove it at inference, so the generation path at inference is unchanged.

%\vspace{-2mm}
\paragraph{Scaffold construction}
Given the support masks $\{A_p\}_{p=1}^{N}$ and reference representations $\{z_p^U\}_{p=1}^{N}$, we first resize each mask to the target resolution and normalize it to $\tilde{A}_p(u) \in [0,1]$, ensuring the peak activation of each mask is scaled to unity.
Each reference representation is projected into a scaffold-specific feature space:
\begin{equation}
    e_p = W_s z_p^U,
\end{equation}
where $W_s$ is a learnable linear projection matrix that aligns the reference representation with the scaffold feature dimension.
The scaffold canvas $C$ at spatial location $u$ is then defined as a weighted sum of these features:
\begin{equation}
    C(u) = \sum_{p=1}^{N} \tilde{A}_p(u) e_p.
    \label{eq:scaffold_const}
\end{equation}
By not normalizing the sum across phrases, $C(u)$ naturally retains the support confidence, where regions with stronger phrase support produce higher-magnitude scaffold features.

%\vspace{-2mm}
\paragraph{Scaffold injection}
The scaffold canvas is injected into an intermediate spatial feature map $F_t^\ell \in \mathbb{R}^{H \times W \times D}$ in the generation path through a lightweight residual adapter:
\begin{equation}
    F_t^\ell(u) \leftarrow F_t^\ell(u) + m \cdot \alpha \cdot W_c C(u),
    \label{eq:scaffold_inj}
\end{equation}
where $W_c$ is a projection layer, $\alpha$ is a learnable scalar initialized to zero, and $m \sim \mathrm{Bernoulli}(\rho)$ is a scaffold dropout variable with the keep probability $\rho \in [0, 1]$.
The residual form exposes the canvas without changing the architecture of the generation path.
Since $\alpha$ is initialized to zero, the injection leaves the forward computation unchanged at the start of training.

We attenuate the injection by decreasing $\rho$ over training with a cosine schedule. 
Because $m$ is sampled per example and independently of the timestep, scaffolded and unscaffolded steps update the same generation parameters. 
As $\rho$ decays, unscaffolded steps dominate, so the parameters shaped with the scaffold converge toward the path used at inference.

\vspace{-0.2cm}
% 4.5 Training Objective
\subsection{Training Objective and Inference}
\label{sec:method-objective}

The training objective comprises the phrase-local contrastive loss and the standard denoising loss:
\begin{equation}
    \mathcal{L} = \mathcal{L}_{\mathrm{gen}} + \lambda \, \mathcal{L}_{\mathrm{phrase}},
    \label{eq:final_obj}
\end{equation}
where $\mathcal{L}_{\mathrm{gen}}$ represents the native denoising objective of the base model and $\lambda$ is a weighting hyperparameter.
During training, the understanding path and the support extraction module remain frozen, while we jointly update the generation path, the projection head $g_\phi$, and the scaffold adapter.
The scaffold does not introduce an explicit loss term; instead, it modulates the training process solely through the forward computation when active.
At inference, both the contrastive objective and the scaffold branch are discarded.
The final model thus operates with its original generation interface and requires no additional inputs or computational overhead during deployment.
Together, these components define our Structural Self-Teaching (SST) framework, which transfers phrase-level structure from the understanding path to the generation path entirely through internal signals.

%-------------------------------------------------------------------
%		Evaluation and Analysis
%-------------------------------------------------------------------
% !TEX root = ../main.tex

\vspace{-0.2cm}
\section{Experiments}
\label{sec:experiments}

\subsection{Setup}
\label{sec:setup}
We train on the Instruction-Following subset of Echo-4o~\cite{ye2025echo} (70K synthetic image-caption pairs) and evaluate on GenEval2~\cite{kamath2025geneval}, which contains 800 prompts stratified by atomicity from 3 to 10 and uses Soft-TIFA, an atom-level 
VQA judge based on Qwen3-VL-8B. 
GenEval2 measures compositional fidelity, that is, whether the image satisfies each atom specified in the prompt.
We report arithmetic mean (AM), geometric mean (GM), category-level scores, and atomicity-stratified GM, using GM as the primary metric since it penalizes missing visual atoms. 
GenEval++ results are reported in Appendix~\ref{app:other-benchmarks}. 
We validate four UMM backbones (BLIP-3o 4B/8B, Show-o2 7B, BAGEL) under LoRA-based tuning of the generation modules with the understanding path frozen. 
Backbone-specific instantiations and training details are provided in Appendix~\ref{app:implementation}.

\begin{table*}[t]
\centering
\caption{
Performance comparison on GenEval2.
We report arithmetic mean (AM), geometric mean (GM), category-level scores, and atomicity-wise GM.
LoRA FT and SST report means over three training runs, with sample standard deviations for AM and GM.
Shaded rows are reference baselines under different settings. 
CoT modifies inference, and RecA tunes the generation parameters directly, without LoRA, with caption-free reconstruction supervision. 
Bold marks the best value in each backbone group.
}
\label{tab:geneval2-main}
\vspace{-1mm}
%\scriptsize
\setlength{\tabcolsep}{2.0pt}
\hspace{-1.5mm}
\scalebox{0.8}{
\begin{tabular}{@{}lcc@{\hspace{1.0em}}ccccc@{\hspace{1.0em}}cccccccc@{}}
\toprule
\multirow{2}{*}{Method}
& \multirow{2}{*}{AM}
& \multirow{2}{*}{GM}
& \multicolumn{5}{c}{Category}
& \multicolumn{8}{c}{Atomicity} \\
\cmidrule(lr){4-8}\cmidrule(lr){9-16}
& & & Obj. & Attr. & Count & Pos. & Verb
& 3 & 4 & 5 & 6 & 7 & 8 & 9 & 10 \\

\midrule
BLIP-3o 4B
& 53.87 & 11.78
& 77.94 & 52.87 & 30.72 & 35.06 & 15.44
& 34.78 & 28.21 & 14.22 & 6.14 & 3.84 & 3.53 & 2.25 & 1.26 \\
+ LoRA FT
& 54.70\std{0.18} & 13.00\std{0.35} 
& 76.62 & 54.24 & 31.64 & 36.33 & 13.37 
& 39.34 & 28.24 & 14.65 & 10.71 & 4.33 & 3.46 & 1.90 & 1.44 \\
+ SST (Ours)
& \textbf{58.29}\std{0.15} & \textbf{16.13}\std{0.28} 
& \textbf{81.29} & \textbf{54.49} & \textbf{37.17} & \textbf{41.53} & \textbf{17.03} 
& \textbf{47.41} & \textbf{33.19} & \textbf{16.92} & \textbf{14.43} & \textbf{5.72} & \textbf{4.53} & \textbf{3.93} & \textbf{2.92} \\

\midrule
BLIP-3o 8B
& 59.35 & 13.50
& 85.08 & 57.79 & 33.46 & 47.13 & 20.34
& 31.63 & 30.90 & 15.59 & 12.30 & 6.79 & 5.61 & 2.67 & 2.52 \\
+ LoRA FT
& 60.48\std{0.23} & 14.86\std{0.48} 
& 86.31 & 58.00 & 34.84 & 46.94 & \textbf{21.22} 
& 33.05 & 32.77 & 16.25 & 13.87 & 7.49 & \textbf{6.92} & \textbf{5.23} & 3.27 \\
+ SST (Ours)
& \textbf{61.82}\std{0.45} & \textbf{16.47}\std{0.39} 
& \textbf{87.88} & \textbf{58.86} & \textbf{35.22} & \textbf{50.17} & 17.14 
& \textbf{35.78} & \textbf{35.92} & \textbf{19.84} & \textbf{15.83} & \textbf{8.63} & 6.88 & 5.14 & \textbf{3.77} \\

\midrule
Show-o2 7B
& 63.43 & 15.32
& 88.36 & 60.51 & 44.01 & 44.05 & \textbf{27.58}
& 31.99 & 33.44 & 16.01 & 17.09 & 7.52 & \textbf{7.69} & 5.74 & 3.00 \\
+ LoRA FT
& 63.42\std{0.38} & 16.03\std{0.41} 
& 88.52 & 61.34 & 43.91 & 43.44 & 18.12 
& 37.44 & 32.22 & 18.82 & 16.66 & \textbf{8.85} & 6.02 & 5.33 & 2.88 \\
+ SST (Ours)
& \textbf{64.28}\std{0.32} & \textbf{17.11}\std{0.58} 
& \textbf{89.44} & \textbf{62.71} & \textbf{44.12} & \textbf{46.27} & 20.89 
& \textbf{39.36} & \textbf{34.67} & \textbf{19.81} & \textbf{17.82} & 8.45 & 7.56 & \textbf{5.86} & \textbf{3.31} \\

\midrule
BAGEL (14B)
& 66.36 & 19.65
& 89.27 & 61.90 & 49.73 & 44.56 & 22.96
& 51.48 & 32.94 & 21.85 & 20.51 & 10.51 & 10.75 & 5.93 & 3.18 \\
\rowcolor{gray!10}
+ CoT
& 70.81 & 23.29
& 91.97 & \textbf{71.70} & 53.07 & 51.51 & 17.99
& 51.94 & 41.58 & 30.74 & 25.55 & 11.99 & 10.97 & 7.45 & 6.10 \\
\rowcolor{gray!10}
+ RecA %\({}^{\dagger}\)
& 72.66 & 25.52
& \textbf{94.21} & 71.63 & 55.65 & 51.55 & \textbf{27.34}
& 58.89 & 44.04 & 28.71 & \textbf{30.20} & 15.12 & 10.84 & 7.77 & \textbf{8.58} \\
+ LoRA FT
& 73.01\std{0.31} & 26.47\std{0.21} 
& 93.23 & 68.43 & 57.39 & 52.71 & 25.32 
& 59.89 & 48.71 & 30.74 & 28.86 & \textbf{15.89} & 11.50 & 8.93 & 7.20 \\

+ SST (Ours)
& \textbf{74.06}\std{0.25} & \textbf{27.33}\std{0.42} 
& 93.39 & 71.69 & \textbf{58.23} & \textbf{54.21} & 26.82
& \textbf{60.52} & \textbf{52.16} & \textbf{31.86} & 28.74 & 14.89 & \textbf{13.72} & \textbf{9.33} & 7.44 \\

\bottomrule
\end{tabular}}
\vspace{-0.4cm}
\end{table*}

\subsection{Quantitative Results}
\label{sec:quantitative-results}
We compare SST with the base UMMs and with LoRA FT, which applies the same LoRA tuning on the same data with the native denoising loss alone (Table~\ref{tab:geneval2-main}).
For BAGEL, we also report CoT prompting and RecA as reference baselines, since they differ from our setting in inference procedure or tuning protocol.

The gains of SST come from its structural supervision rather than from the training data.
LoRA FT trains on the same data, and SST improves over it in AM and GM on all four backbones, which span three generation architectures.
The gains exceed the standard deviation across three training runs.
On BLIP-3o 4B and BAGEL, paired bootstrap intervals over prompts exclude zero (Appendix~\ref{app:bootstrap}), and a blinded human evaluation also favors SST (Appendix~\ref{app:human_eval}).

The category-level scores follow what the supports cover.
The supports localize noun phrases, and SST scores higher than LoRA FT on Object, Attribute, Count, and Position on all four backbones. 
Verb, which the supports do not encode and only 82 of the 800 prompts contain, is the exception.
On Show-o2, LoRA fine-tuning lowers Verb sharply and SST recovers only part of the drop; on BLIP-3o 8B, SST falls below both baselines.
On BAGEL, SST exceeds both reference baselines in AM and GM without changing the inference procedure, while RecA leads on Object and Verb.

\subsection{Ablation Studies and Analysis}
\label{sec:ablation}
We analyze the contribution of each component, the source of the support masks, and the design choices of the scaffold, and we examine whether SST reduces generation diversity.
Table~\ref{tab:component-ablation} reports means over three training runs on BLIP-3o 4B and BAGEL. 
Unless noted otherwise, the remaining ablations use a single run on BLIP-3o 4B, with the same training data, number of optimization steps, and evaluation protocol as the main experiments.

\begin{table}[t]
\centering
\caption{Results of main ablation studies on the phrase-local contrastive alignment ($\mathcal{L}_{phrase}$) and structural scaffold on BLIP-3o 4B and BAGEL. 
Means and sample standard deviations over three training runs. 
Bold marks the best value in each column for each backbone.}
\vspace{0.2cm}
\label{tab:component-ablation}
\setlength{\tabcolsep}{2pt}
\scalebox{0.8}{%
\begin{tabular}{llccccccc}
\toprule
 & Method & AM & GM & Object & Attribute & Count & Position & Verb \\
\midrule
\multirow{4}{*}{BLIP-3o 4B}
 & LoRA FT          & 54.70\std{0.18} & 13.00\std{0.35} & 76.62\std{0.42} & 54.24\std{0.68} & 31.64\std{0.44} & 36.33\std{0.59} & 13.37\std{0.66} \\
 & Phrase-local only & 56.04\std{0.34} & 15.42\std{0.41} & 79.88\std{0.52} & 54.23\std{0.37} & 35.77\std{0.66} & 37.98\std{0.31} & \textbf{17.28}\std{0.84} \\
 & Scaffold only    & 55.24\std{0.28} & 14.43\std{0.33} & 77.42\std{0.34} & 53.94\std{0.24} & 33.12\std{0.56} & 38.67\std{0.44} & 14.42\std{0.83} \\
 & Full SST         & \textbf{58.29}\std{0.15} & \textbf{16.13}\std{0.28} & \textbf{81.29}\std{0.25} & \textbf{54.49}\std{0.20} & \textbf{37.17}\std{0.59} & \textbf{41.53}\std{0.42} & 17.03\std{0.78} \\
\midrule
\multirow{4}{*}{BAGEL}
 & LoRA FT          & 73.01\std{0.31} & 26.47\std{0.21} & 93.23\std{0.41} & 68.43\std{0.49} & 57.39\std{1.27} & 52.71\std{0.55} & 25.32\std{1.09} \\
 & Phrase-local only & 73.69\std{0.30} & 27.12\std{0.39} & \textbf{93.47}\std{0.35} & 70.24\std{0.58} & 58.20\std{0.41} & 52.78\std{0.36} & 26.47\std{0.71} \\
 & Scaffold only    & 73.44\std{0.32} & 26.79\std{0.37} & 93.33\std{0.22} & 69.98\std{0.24} & \textbf{58.42}\std{0.66} & 53.92\std{0.38} & 26.14\std{0.62} \\
 & Full SST         & \textbf{74.06}\std{0.25} & \textbf{27.33}\std{0.42} & 93.39\std{0.28} & \textbf{71.69}\std{0.76} & 58.23\std{0.74} & \textbf{54.21}\std{0.51} & \textbf{26.82}\std{0.67} \\
\bottomrule
\end{tabular}}
\vspace{-0.3cm}
\end{table}

\iffalse
\begin{table}[t]
\centering
\caption{
Results of main ablation studies on the phrase-local contrastive alignment ($\mathcal{L}_{phrase}$) and structural scaffold. 
Bold numbers indicate the best performance in each column.
}
\vspace{0.2cm}
\label{tab:component-ablation}
\setlength{\tabcolsep}{12pt}
\scalebox{0.8}{
\begin{tabular}{cccccccc}
\toprule
$\mathcal{L}_{phrase}$ & Scaffold & Object & Attribute & Count & Position &  Verb & Overall (GM) \\
\midrule
&& 76.22 & 53.60 & 31.40 & 36.29 & 13.60 & 13.35 \\
$\surd$ &  & 79.37 & 53.86 & 36.20 & 38.27 & \textbf{18.16} & 15.48 \\
& $\surd$ & 77.12 & 53.78 & 32.94 & 38.54 & 13.48 & 14.23 \\
$\surd$ & $\surd$ & \textbf{81.43} & \textbf{54.34} & \textbf{37.32} & \textbf{41.57} & 16.29 & \textbf{16.05} \\
\bottomrule
\end{tabular}
}
\vspace{-0.3cm}
\end{table}
\fi

\paragraph{Effect of each component}
Table~\ref{tab:component-ablation} ablates Phrase-Local Contrastive Alignment and the Structural Scaffold on BLIP-3o 4B and BAGEL.
Phrase-local alignment improves GM over LoRA FT on both backbones, and the full model achieves the highest GM.
The two mechanisms overlap in the categories they improve but differ in where their gains concentrate.
On BLIP-3o 4B, phrase-local alignment accounts for most of the gain on Object, Count, and Verb.
The scaffold contributes most clearly on Position: scaffold only ranks below phrase-local only in GM on both backbones, yet above it on Position. 
Adding the scaffold to phrase-local alignment raises Position beyond run-to-run variation on both backbones, and on BLIP-3o 4B it also raises Object and Count. 
On BAGEL, phrase-local alignment alone leaves Position at the level of LoRA FT.

\begin{table}[t]
\centering
\caption{Mask source on BLIP-3o 4B (single run each). 
Grounded-SAM masks serve as an external reference that replaces the attention-derived supports during training, with all other settings fixed. 
The last row reports the difference between the attention-derived supports and the external reference.}
\vspace{0.2cm}
\label{tab:mask_source}
\scalebox{0.8}{
\begin{tabular}{lccccccc}
\toprule
Mask source & AM & GM & Object & Attribute & Count & Position & Verb \\
\midrule
None (LoRA FT)                     & 54.56 & 13.35 & 76.22 & 53.60 & 31.40 & 36.29 & 13.60 \\
\midrule
Grounded-SAM (external)          & 59.58 & 16.84 & 83.09 & 55.68 & 35.28 & 40.52 & 18.64 \\
Attention-derived (SST)            & 58.34 & 16.05 & 81.43 & 54.34 & 37.32 & 41.57 & 16.29 \\
\midrule
$\Delta$ (SST $-$ external)        & $-1.24$ & $-0.79$ & $-1.66$ & $-1.34$ & $+2.04$ & $+1.05$ & $-2.35$ \\
\bottomrule
\end{tabular}}
\end{table}

\vspace{-0.2cm}
\paragraph{Support masks}
Table~\ref{tab:mask_source} replaces the attention-derived supports with masks from an external segmentation model, Grounded-SAM, keeping all other settings fixed. 
The attention-derived supports come within 0.79 GM of the external masks and achieve 77\% of their gain over LoRA FT.  
Coarse supports from the understanding path suffice for SST, without an external segmentation model.

\begin{table}[t]
\centering
\caption{
Ablation on $\lambda$, the weighting of $\mathcal{L}_{\text{phrase}}$. 
The optimal scale of $\lambda$ depends on the backbone. 
We report Overall GM (\%) on GenEval2 and bold the best $\lambda$ per backbone (single run each). %use the best $\lambda$ per backbone (in bold) in all main experiments.
}
\vspace{0.2cm}
\label{tab:lambda_ablation}
\scalebox{0.8}{
\begin{tabular}{l c c c c c l c c c c}
%\toprule
\multicolumn{5}{c}{BLIP-3o variants} & & \multicolumn{5}{c}{Show-o2 / BAGEL} \\
\cmidrule(lr){1-5} \cmidrule(lr){7-11}
$\lambda$  & 1   & 5   & 10  & 15  & & $\lambda$  & 0.01 & 0.1  & 0.2 & 0.5 \\
\midrule
BLIP-3o 4B & 14.11 & 15.36 & \textbf{16.05} & 14.71 & & Show-o2 7B & 15.66 & \textbf{16.55} & 16.12 & 15.44 \\
BLIP-3o 8B & 14.55 & 16.18 & \textbf{16.89} & 15.23 & & BAGEL      & 26.14 & \textbf{27.21} & 26.40 & 25.76 \\
\bottomrule
\end{tabular}
}
\vspace{-0.3cm}
\end{table}

%\vspace{-0.2cm}
\paragraph{Loss weighting}
We ablate $\lambda$, the weighting of $\mathcal{L}_{\text{phrase}}$. 
The optimal scale differs across backbones, with BLIP-3o variants preferring $\lambda \sim 10$ and Show-o2 and BAGEL preferring $\lambda \sim 0.1$ (Table~\ref{tab:lambda_ablation}).
The split follows the generation target space: BLIP-3o denoises in the EVA-CLIP feature space, whereas Show-o2 and BAGEL predict velocities in a VAE latent space.
We use the best $\lambda$ per backbone in all main experiments.

\begin{table}[t]
\centering
\caption{
Scaffold design ablations on GenEval2.
We evaluate scaffold injection layer and dropout schedule while keeping all other settings fixed.
Since the scaffold modulates spatial generation states, we report Position and Overall GM.
Bold numbers indicate the best performance in each column.
}
\label{tab:scaffold-design}
\vspace{0.2cm}
\hspace*{-0.04\linewidth}
\begin{minipage}[t]{0.45\linewidth}
\centering
(a) Scaffold injection layer\par%\vspace{0.3em}
\scalebox{0.8}{
\begin{tabular}{lcc}
\toprule
Layer & Position & Overall (GM) \\
\midrule
No scaffold & 38.27 & 15.48 \\ \midrule
0  & \textbf{41.57} & \textbf{16.05} \\
3  & 40.79 &  15.86 \\
7  & 40.05 & 15.45 \\
11 & 39.46 & 15.52 \\
\bottomrule
\end{tabular}
}
\end{minipage}
%\hfill
\begin{minipage}[t]{0.48\linewidth}
\centering
(b) Scaffold dropout schedule\par%\vspace{0.3em}
\scalebox{0.8}{
\begin{tabular}{lcc}
\toprule
Schedule & Position & Overall (GM) \\
\midrule
No scaffold & 38.27 & 15.48 \\ \midrule
Fixed \(\rho=1.0\) & 38.14 & 14.28 \\
Fixed \(\rho=0.5\) & 40.15 & 15.76 \\
Fixed \(\rho=0.25\) & 39.24 & 15.32  \\
Decay of \(\rho\) & \textbf{41.57} & \textbf{16.05} \\
\bottomrule
\end{tabular}
}
\end{minipage}
%\vspace{-3mm}
\end{table}

\vspace{-2mm}
\paragraph{Scaffold injection layer}
Table~\ref{tab:scaffold-design}(a) evaluates different injection layers for the structural scaffold.
Since the scaffold modulates spatial generation states, we report Position as the most directly relevant category together with Overall GM.
The earliest layer performs best, and Position decreases as the scaffold is moved to later layers, suggesting that scaffold modulation is most effective before the generation path has committed to a spatial layout within each forward pass.

\vspace{-2mm}
\paragraph{Scaffold scheduling}
Table~\ref{tab:scaffold-design}(b) compares scaffold dropout schedules.
Since the scaffold branch is removed at inference, the schedule determines how much the generation path is exposed to the privileged scaffold signal during training.
The decay schedule achieves the best Position and Overall GM, outperforming fixed keep probabilities.
This suggests that the scaffold should provide stronger guidance early in training and be gradually phased out as the unscaffolded generation path is optimized.
The result supports scaffold dropout as a training-time curriculum rather than a persistent conditioning pathway.

\vspace{-2mm}
\paragraph{Scaffold mechanism}
\begin{wraptable}{r}{0.30\textwidth}
\vspace{-1.5em}
\centering
\small
\caption{Canvas controls.
Uniform removes structure; 
Shuffled randomizes binding.
}
\vspace{-0.5em}
\label{tab:canvas}
\begin{tabular}{l c}
\toprule
Canvas & GM \\
\midrule
No scaffold & 15.48 \\
Uniform & 15.37 \\
Shuffled & 14.69 \\
Full (Ours) & \textbf{16.05} \\
\bottomrule
\end{tabular}
\vspace{-1em}
\end{wraptable}
We isolate the source of the scaffold benefit through canvas controls. 
Replacing the phrase-grounded canvas with a Uniform canvas, which removes spatial structure, yields performance comparable to no-scaffold, while a Shuffled canvas, which preserves spatial structure but randomizes phrase--region binding, falls below it. 
The full canvas outperforms both controls (Table~\ref{tab:canvas}), indicating that the benefit derives from correctly binding phrase content to spatial support, rather than from added conditioning capacity alone.

\newcommand{\ci}[3]{#1$_{[#2,\,#3]}$}  % 프리앰블 또는 표 바로 앞

\begin{table}[t]
\centering
\caption{Generation diversity on GenEval2 (800 prompts). 
For each prompt, we generate four samples and average the distance over all six pairs: LPIPS, and the cosine distance between DINOv2 features. 
Higher is more diverse. 
Subscripts are 95\% bootstrap intervals over prompts.}
\label{tab:diversity}
\vspace{0.2cm}
\scalebox{0.8}{
\setlength{\tabcolsep}{4pt}
\begin{tabular}{lcccc}
\toprule
 & \multicolumn{2}{c}{BLIP-3o 4B} & \multicolumn{2}{c}{BAGEL} \\
\cmidrule(lr){2-3} \cmidrule(lr){4-5}
Method & LPIPS & DINOv2 & LPIPS & DINOv2 \\
\midrule
Base       & \ci{0.321}{0.316}{0.326} & \ci{0.127}{0.121}{0.133} & \ci{0.586}{0.579}{0.592} & \ci{0.300}{0.293}{0.308} \\
LoRA FT    & \ci{0.283}{0.275}{0.291} & \ci{0.098}{0.093}{0.102} & \ci{0.628}{0.618}{0.638} & \ci{0.243}{0.235}{0.251} \\
SST (Ours) & \ci{0.317}{0.308}{0.326} & \ci{0.123}{0.117}{0.128} & \ci{0.626}{0.614}{0.638} & \ci{0.256}{0.249}{0.263} \\
\bottomrule
\end{tabular}
}
\end{table}

\vspace{-0.2cm}
\paragraph{Generation diversity}
Table~\ref{tab:diversity} measures generation diversity among four samples per GenEval2 prompt with two distances, LPIPS~\cite{zhang2018lpips} and the cosine distance between DINOv2~\cite{oquab2024dinov2} features.
Diversity depends mainly on the backbone, where BAGEL yields roughly twice the distance of BLIP-3o 4B under both metrics. 
Within each backbone, SST yields higher diversity than LoRA FT under both metrics on BLIP-3o 4B. 
On BAGEL, LPIPS is similar between the two methods, while DINOv2 distance is higher with SST.

\begin{figure}[t]
\centering
\setlength{\tabcolsep}{2pt}
\renewcommand{\arraystretch}{1.0}

% Row 1: prompts 1, 2
\begin{tabular}{@{}
>{\centering\arraybackslash}m{0.15\textwidth}
>{\centering\arraybackslash}m{0.15\textwidth}
>{\centering\arraybackslash}m{0.15\textwidth}
@{\hspace{10pt}}
>{\centering\arraybackslash}m{0.15\textwidth}
>{\centering\arraybackslash}m{0.15\textwidth}
>{\centering\arraybackslash}m{0.15\textwidth}
@{}}
\multicolumn{3}{c}{\shortstack{\sffamily\footnotesize\textit{BLIP-3o 4B} : \sffamily\footnotesize ``a penguin and three pink cows''}}
& 
\multicolumn{3}{c}{\shortstack{\sffamily\footnotesize\textit{BLIP-3o 8B} : ``a candle to the left of a brown kangaroo \\ \sffamily\footnotesize to the left of a toy''}}
\\[2pt]
{\sffamily\scriptsize Base} & {\sffamily\scriptsize LoRA FT} & {\sffamily\scriptsize SST (Ours)} & {\sffamily\scriptsize Base} & {\sffamily\scriptsize LoRA FT} & {\sffamily\scriptsize SST (Ours)}
\\
\includegraphics[width=0.15\textwidth]{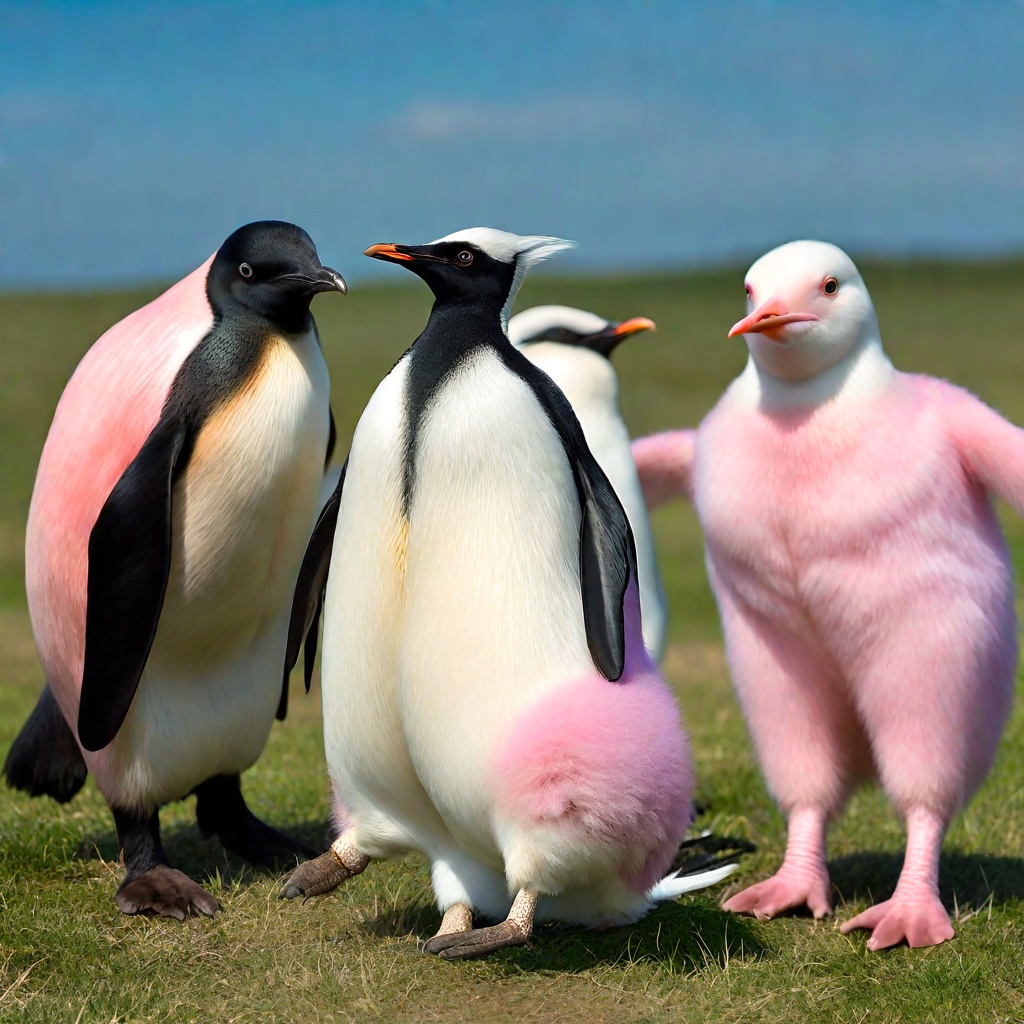}
& \includegraphics[width=0.15\textwidth]{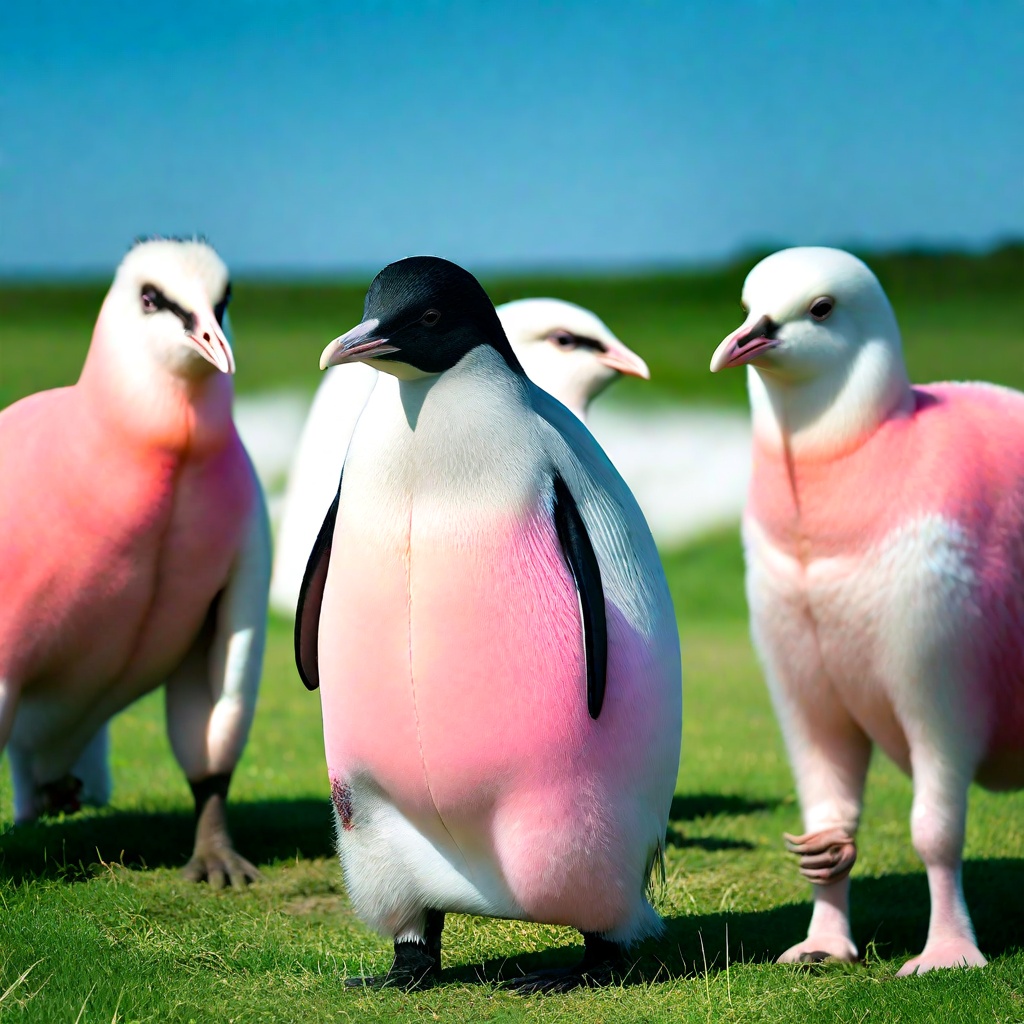}
& \includegraphics[width=0.15\textwidth]{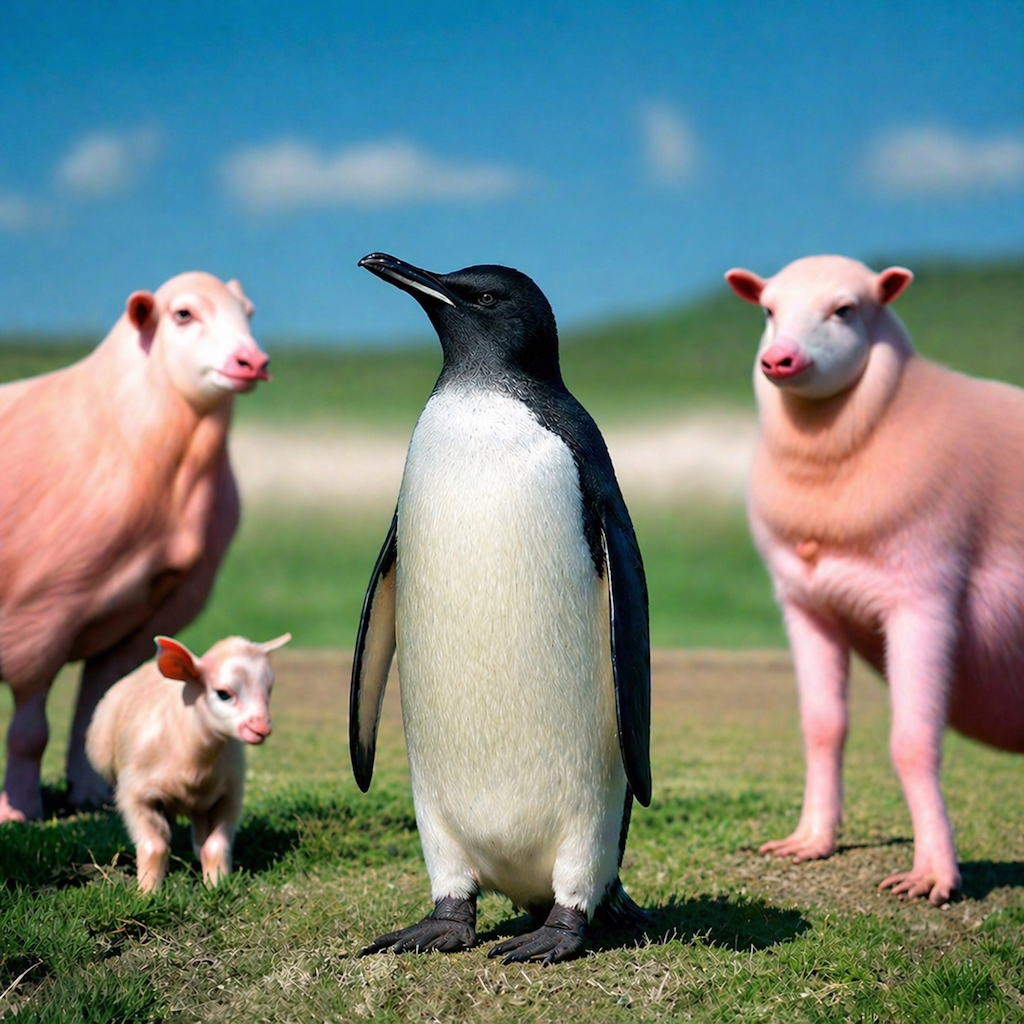}
& \includegraphics[width=0.15\textwidth]{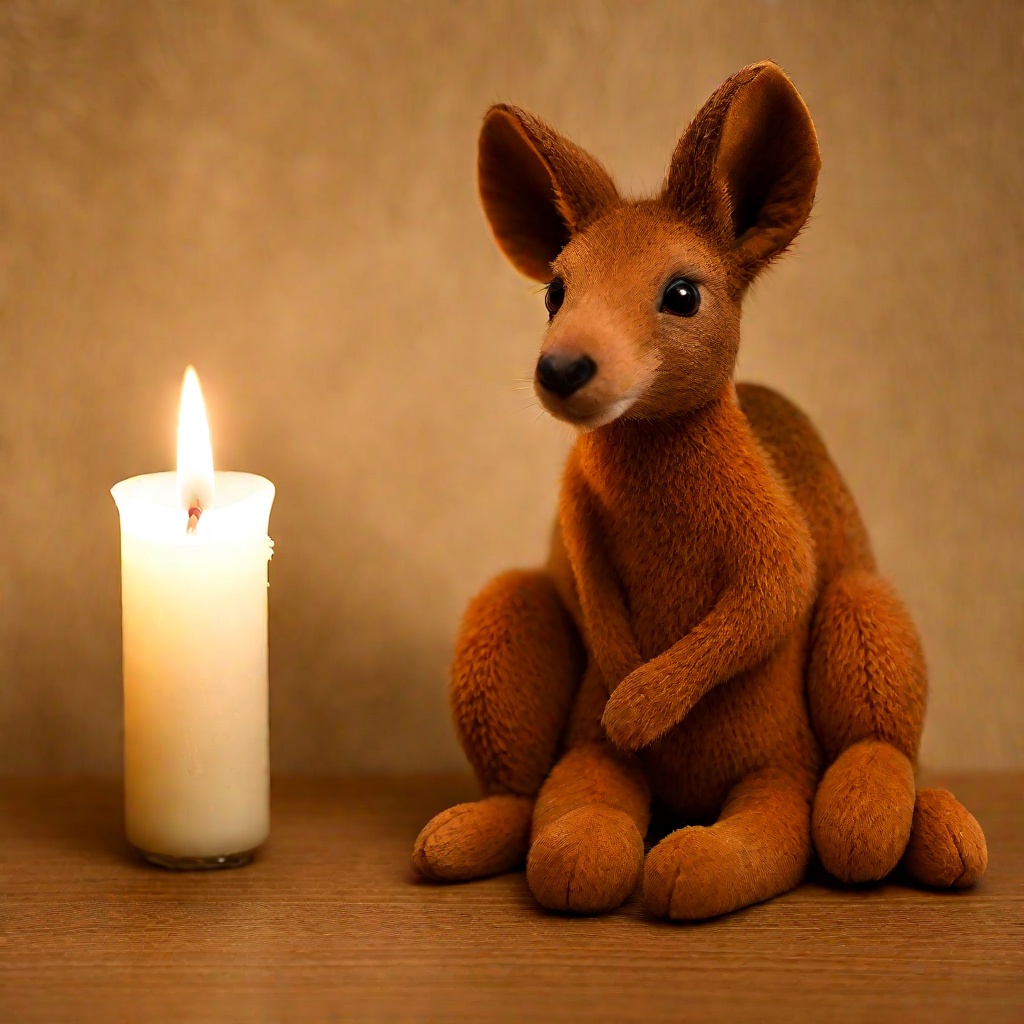}
& \includegraphics[width=0.15\textwidth]{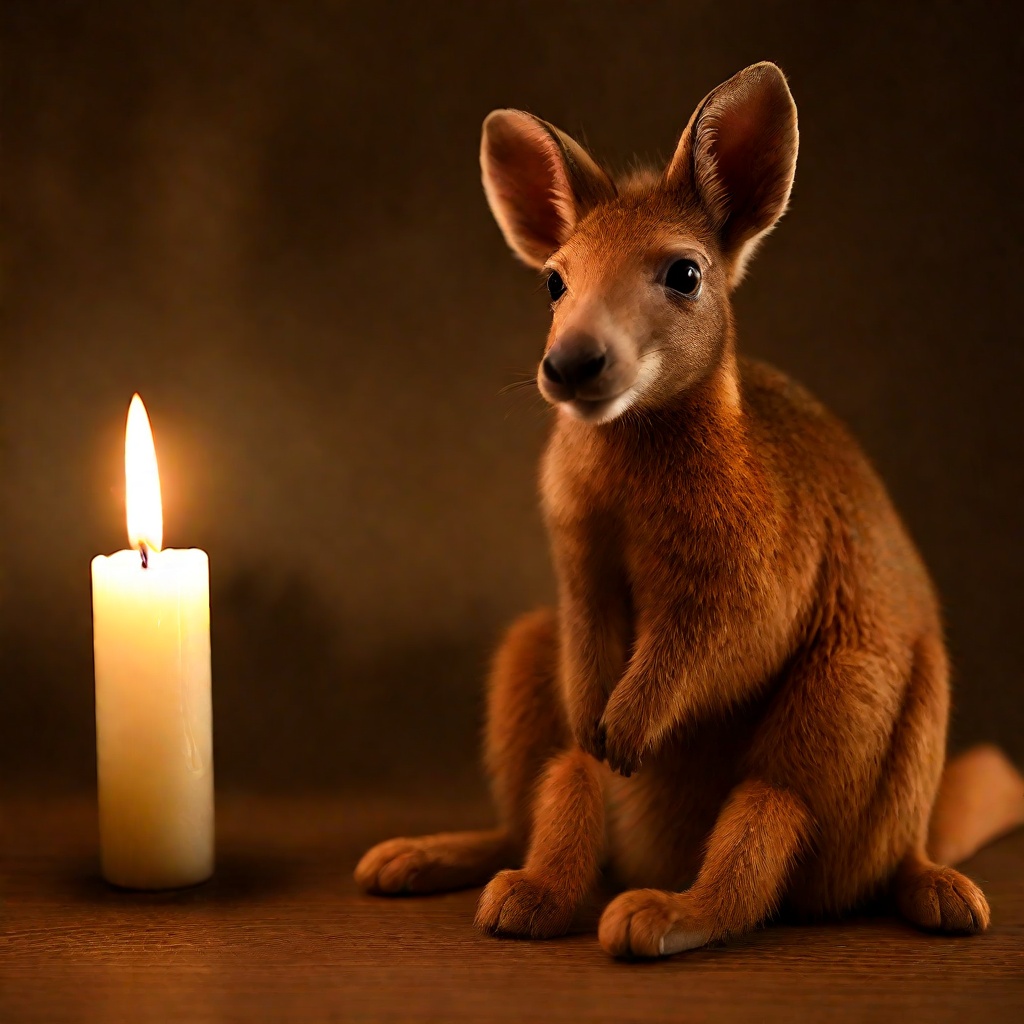}
& \includegraphics[width=0.15\textwidth]{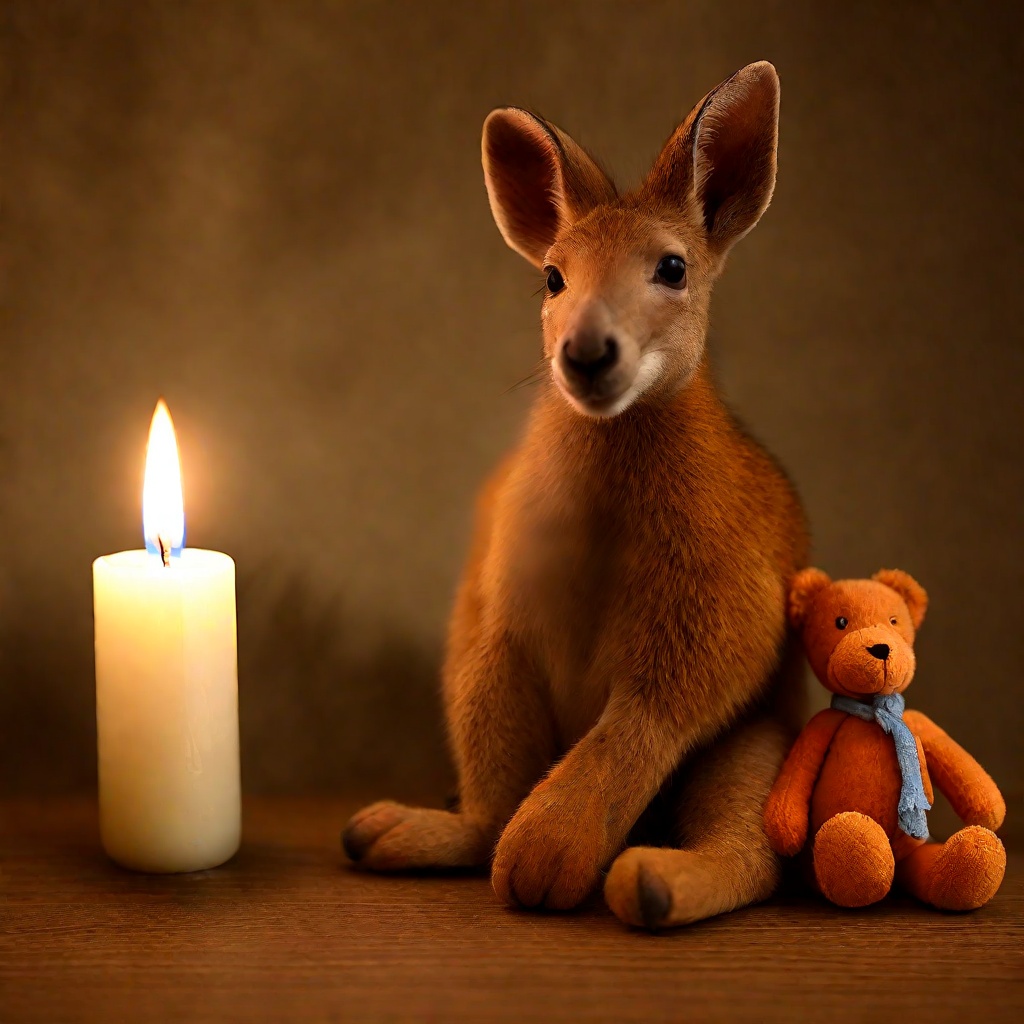}
\\\noalign{\vspace{10pt}} %[6pt]

% Row 2: prompts 3, 4
\multicolumn{3}{c}{\shortstack{\sffamily\footnotesize\textit{Show-o2 7B} : \sffamily\footnotesize ``a white violin and a red bicycle''}}
&
\multicolumn{3}{c}{\shortstack{\sffamily\footnotesize\textit{BAGEL} : \sffamily\footnotesize ``a metal clock on top of a bear''}}
\\[2pt]
{\sffamily\scriptsize Base} & {\sffamily\scriptsize LoRA FT} & {\sffamily\scriptsize SST (Ours)} & {\sffamily\scriptsize Base} & {\sffamily\scriptsize LoRA FT} & {\sffamily\scriptsize SST (Ours)}
\\
\includegraphics[width=0.15\textwidth]{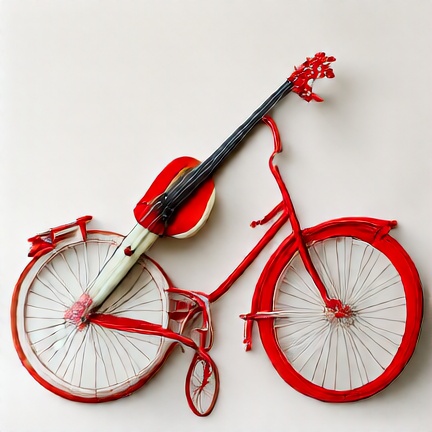}
& \includegraphics[width=0.15\textwidth]{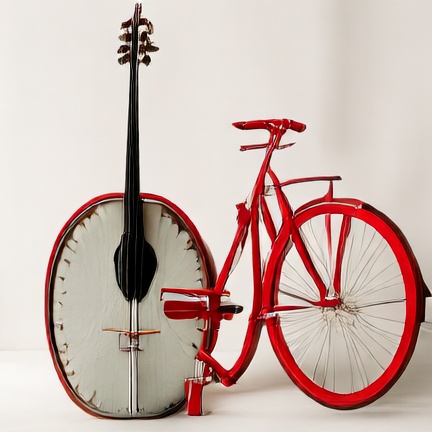}
& \includegraphics[width=0.15\textwidth]{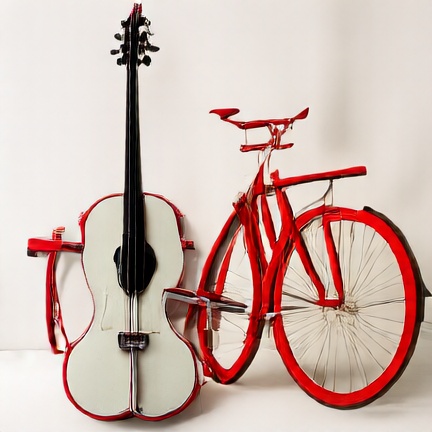}
& \includegraphics[width=0.15\textwidth]{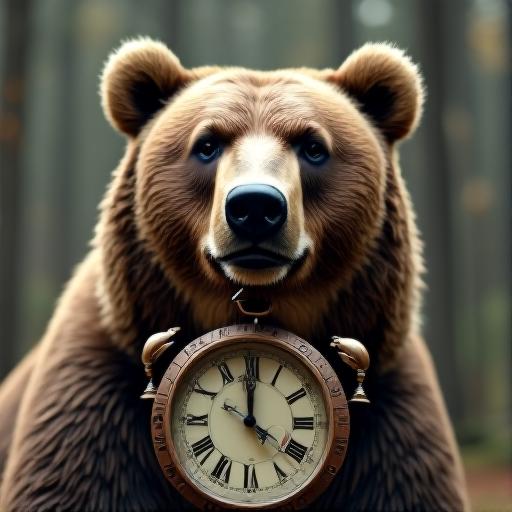}
& \includegraphics[width=0.15\textwidth]{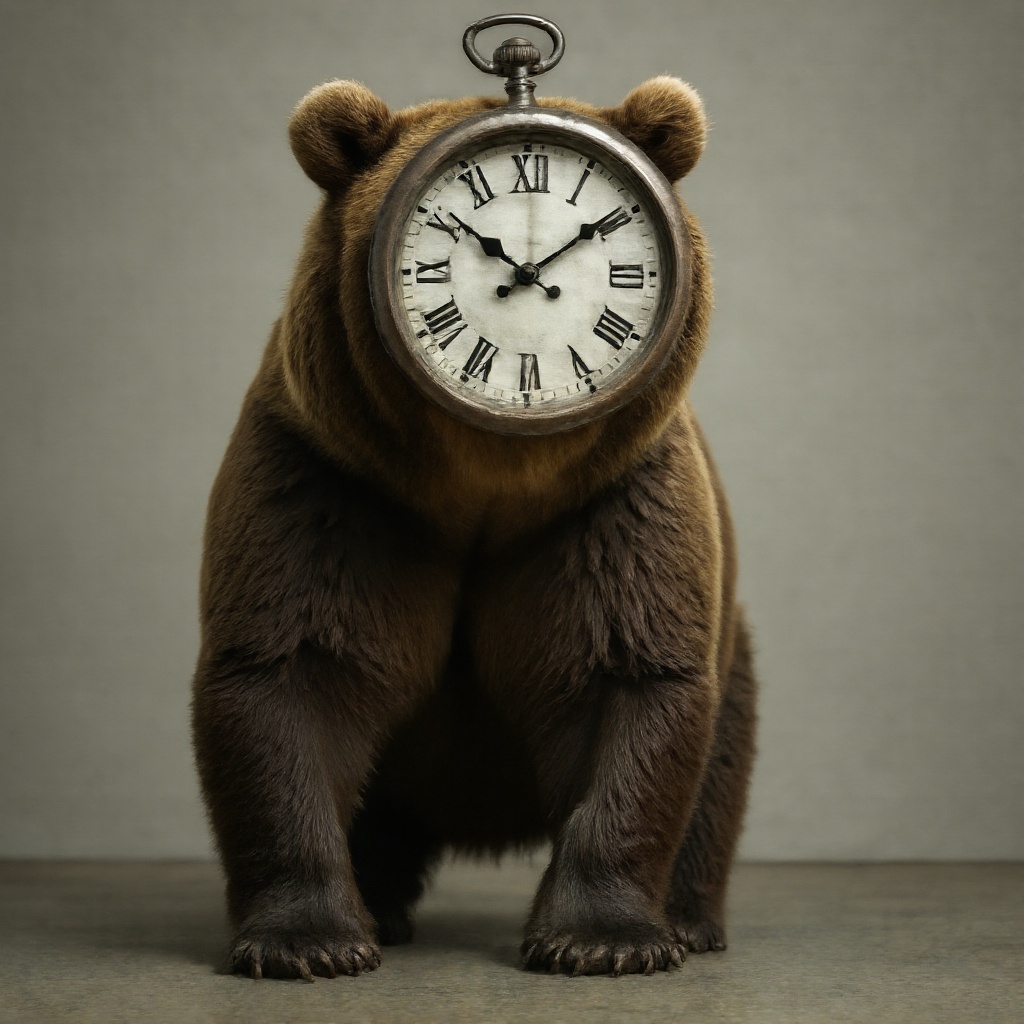}
& \includegraphics[width=0.15\textwidth]{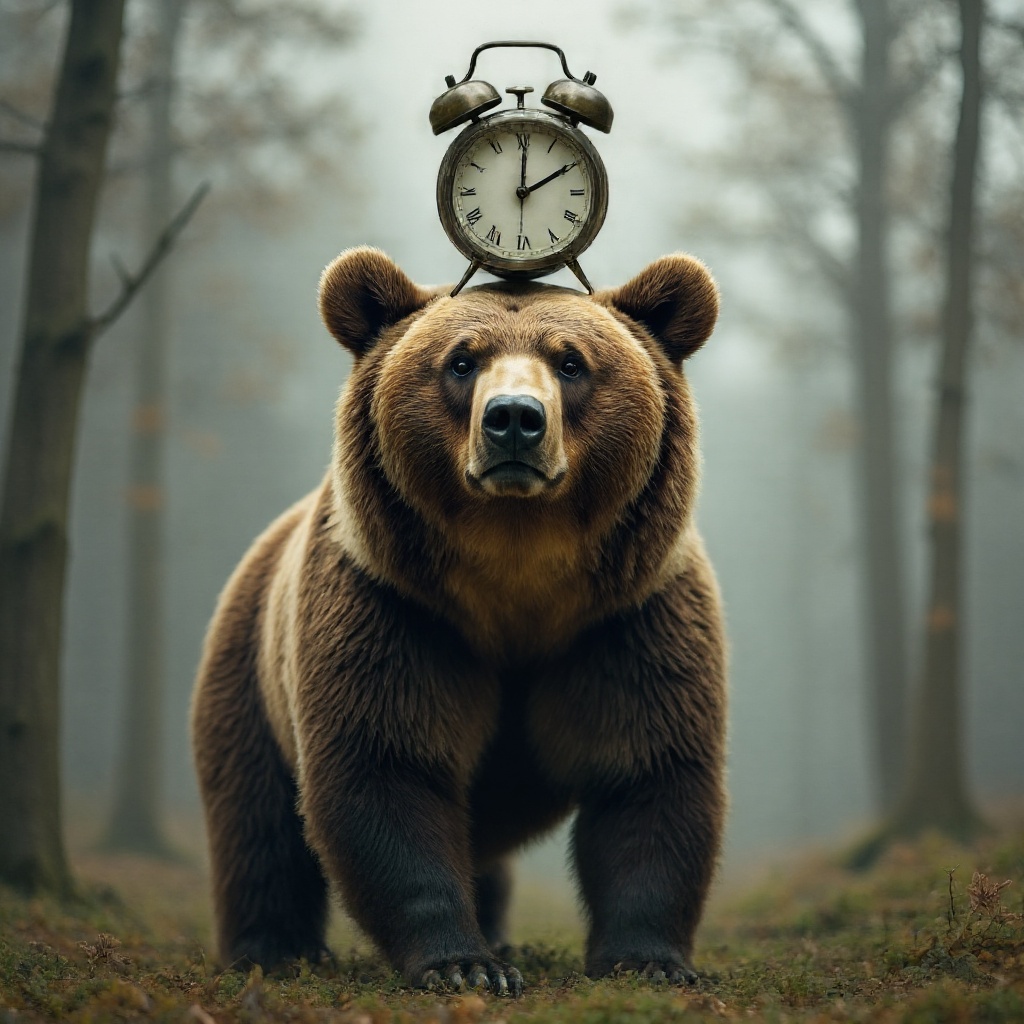}
\\
\end{tabular}
\caption{
Qualitative comparison across four UMM backbones. 
Each group of three images shows one failure mode of the base model: 
attribute mis-binding for BLIP-3o 4B (cows fused into penguin-like hybrids), 
object omission for BLIP-3o 8B (the toy is missing), 
object entanglement for Show-o2 7B (violin and bicycle fused; LoRA FT also fails to fully separate them), 
and spatial violation for BAGEL (the clock is held rather than placed on top of the bear). 
SST restores the prompt-specified compositional structure across all cases, while LoRA fine-tuning provides only partial mitigation.
}
\label{fig:qualitative}
\end{figure}

\subsection{Qualitative Results}
Figure~\ref{fig:qualitative} presents qualitative comparisons across the four UMM backbones, illustrating how SST resolves distinct compositional failure modes of the base UMMs. 
The selected cases span attribute mis-binding, object omission, object entanglement, and spatial violation, covering the major error types observed in dense compositional prompts. 
Across all cases, the base UMMs either drop, fuse, or mis-relate the prompt-specified concepts, and LoRA fine-tuning provides only partial mitigation while preserving similar failure structures. 
SST instead restores the prompt-specified compositional structure, generating distinct objects with correct attributes and spatial relations. 
These cases illustrate the failure modes that SST reduces: omission, fusion, and misplacement of prompt-specified concepts.
% \vspace{-0.2cm}

%-------------------------------------------------------------------
%		Conclusion
%-------------------------------------------------------------------
% !TEX root = ../main.tex

\section{Conclusion}
\label{sec:conclusion}
We present Structural Self-Teaching (SST), a framework that improves compositional generation in diffusion-based unified multimodal models. 
SST leverages the model's own understanding path as an internal source of phrase-level structural supervision, converting phrase-conditioned attention into support masks without requiring external annotations.
The masks reach the generation path through two complementary mechanisms.
Phrase-local contrastive alignment aligns each generated phrase region with its target region and separates distinct phrases within a prompt. 
A training-only structural scaffold modulates an intermediate spatial state with a phrase-level layout built from the same masks.
Experiments across representative UMMs demonstrate the effectiveness of SST, improving compositional fidelity across object, attribute, count, and spatial constraints.

\clearpage
\section*{Acknowledgement}
% NPRC: Samsung Electronics Co., Ltd. (IO250418-12669-01)
% IITP
%   - XAI: No.RS-2022-II220959 (No.2022-0-00959), (Part 2) Few-Shot Learning of Causal Inference in Vision and Language for Decision Making
%   - AI Grad school: No.RS-2021-II211343, Artificial Intelligence Graduate School Program (Seoul National University)
%   - AI STAR Fellowship: RS-2025-25442338, AI star Fellowship Support Program (Seoul National University)
% NRF
%   - CRC: No.RS-2022-NR070855, Trustworthy Artificial Intelligence
This work was partly supported by Samsung Electronics Co., Ltd. (IO250418-12669-01), and the Institute of Information \& communications Technology Planning \& Evaluation (IITP) grants [No.RS-2022-II220959 (No.2022-0-00959), (Part 2) Few-Shot Learning of Causal Inference in Vision and Language for Decision Making, No.RS-2021-II211343, Artificial Intelligence Graduate School Program (Seoul National University), RS-2025-25442338, AI star Fellowship Support Program (Seoul National University)], and the National Research Foundation of Korea (NRF) grant [No.RS-2022-NR070855, Trustworthy Artificial Intelligence], funded by the Korean government (MSIT).
%\section*{References}

\bibliography{egbib}
\bibliographystyle{splncs}

%%%%%%%%%%%%%%%%%%%%%%%%%%%%%%%%%%%%%%%%%%%%%%%%%%%%%%%%%%%%
\clearpage
\appendix

%-------------------------------------------------------------------
%		Appendix
%-------------------------------------------------------------------
% !TEX root = ../main.tex

\appendix

%========================================================
\section{Implementation Details}
\label{app:implementation}

\subsection{Training Configuration}
\label{app:training}
Table~\ref{tab:training_config} summarizes the training configuration. 
All backbones are trained for 5{,}000 steps with a per-device batch size of 1 across 4 GPUs and 8 gradient accumulation steps, yielding an effective batch size of 32. 
We use the AdamW optimizer with 200 warmup steps. 
LoRA rank, alpha, and learning rate vary across backbones as listed below; all other hyperparameters are shared.

\begin{table}[t]
\centering
\caption{Training configuration across backbones.}
\label{tab:training_config}
\scalebox{0.8}{
\begin{tabular}{l c c c}
\toprule
Backbone & LoRA $r$ & LoRA $\alpha$ & Learning rate \\
\midrule
BLIP-3o 4B & 64 & 16 & $1 \times 10^{-4}$ \\
BLIP-3o 8B & 64 & 16 & $5 \times 10^{-5}$ \\
Show-o2 7B & 32 & 16 & $1 \times 10^{-5}$ \\
BAGEL & 64 & 16 & $1 \times 10^{-4}$ \\
\bottomrule
\end{tabular}
}
\end{table}

\subsection{Architecture-Specific Instantiations}
\label{app:backbones}

We instantiate SST on three unified multimodal models: BLIP-3o, Show-o2, and BAGEL. 
Each backbone extracts its own support masks $\{A_p\}$ offline from its understanding path, following the same procedure (Appendix~\ref{app:mask-extraction}).
The canvas construction and the zero-initialized residual adapter of Section~\ref{sec:method-scaffold} are also identical across backbones.

The backbones differ in three choices, each of which follows from the native generation pathway.
The student feature is the generation-side state that we pool for ${z_p^G}$.
The reference feature $v$ is the feature of the target image that we pool for $z^U_p$.
The generation module is the submodule that we tune with LoRA and whose first layer receives the scaffold. Table~\ref{tab:backbones} summarizes these choices. 
In every case, we L2-normalize the pooled representations and apply the objective of Equation~\ref{eq:contrastive_obj}.

\paragraph{BLIP-3o (cascaded DiT)}
BLIP-3o's generation path operates natively in the EVA-CLIP~\cite{sun2023eva} feature space, and the DiT directly outputs $\hat{x}_0$ in that space. 
We attach a projection head $g_\phi$ that maps $\hat{x}_0$ into the EVA-CLIP prototype space used for contrastive alignment.
The DiT cross-attention is tuned with LoRA. 
The scaffold is injected as a residual on the output of the first NextDiT~\cite{zhuo2024lumina} block, whose token grid is already aligned with the latent spatial layout.

\paragraph{Show-o2 (LLM with diffusion head)}
Show-o2’s diffusion head predicts velocities in a VAE latent space. 
We recover $\hat x_0$ from the predicted velocity, patchify it, and attach a linear projection head $g_\phi$ that maps each patch into the space of the reference feature, Show-o2’s unified visual representation of the clean latent of the target image.
The diffusion head is tuned with LoRA and $g_{\phi}$ is trained from scratch. 
The scaffold is injected on the first \texttt{diffusion\_head\_a} block. 
Because Show-o2 packs \texttt{[text\,$|$\,spatial\,$|$\,$\ldots$]} in a single sequence, the injection hook slices the spatial sub-region using the recorded modality positions, applies the adapter, and writes the modified state back into the corresponding positions.

\paragraph{BAGEL (LLM-as-DiT)}
BAGEL has no separate DiT module; the LLM itself predicts velocities for the image latents through a generation-specific MoE branch within each LLM layer. 
We attach a projection head $g_{\phi}$ that maps LLM hidden states to the SigLIP~\cite{zhai2023sigmoid} space, producing the student feature for contrastive alignment. 
We tune the generation MoE of every LLM layer with LoRA, and train $g_{\phi}$ from scratch. 
The scaffold is injected on the first LLM layer. 
Since BAGEL packs \texttt{[text\,$|$\,latent\,$|$\,$\ldots$]} in a single sequence, the injection hook acts only on the latent-token positions and leaves text positions untouched.

\begin{table}[t]
\centering
\small
\caption{Backbone-specific instantiation of phrase-local contrastive alignment and structural scaffold injection. 
All other components, including the support masks, canvas construction, and residual adapter, are shared across backbones.}
\label{tab:backbones}
\scalebox{0.8}{
\begin{tabular}{lccc}
\toprule
                         & BLIP-3o                & Show-o2                       & BAGEL                              \\
\midrule
Student feature  & $\hat x_0$ (EVA-CLIP)       & patchified $\hat x_0$ (VAE latent)         & final LLM hidden state              \\
Reference feature & EVA-CLIP feature       & unified visual representation         & SigLIP feature              \\
Trainable modules (LoRA) & DiT cross-attention    & diffusion head                & generation MoE (all LLM layers)    \\
Scaffold injection point & first NextDiT block    & first \texttt{diffusion\_head\_a} block & first LLM layer (latent positions) \\
\bottomrule
\end{tabular}
}
\end{table}

\subsection{Compute Resources}
\label{app:compute}
% GPU type and count
% Training time per backbone
% Inference cost (unchanged; scaffold removed)
We train BLIP-3o 4B on 4$\times$NVIDIA A5000 GPUs, requiring approximately 11 hours of training. 
The remaining backbones (BLIP-3o 8B, Show-o2 7B, and BAGEL) are trained on 4$\times$NVIDIA A6000 GPUs, each requiring approximately 10 hours.

Support extraction runs once per backbone and dataset. 
On BLIP-3o 4B, it takes about 10 hours on 4 A5000 GPUs for the 70K pairs and produces 4.2 GB of cached supports, which all training runs reuse. 
Extraction materializes attention maps, which precludes fused attention kernels such as FlashAttention. 
During training, the frozen encoder computes the reference features without gradients, and on BLIP-3o the denoising target already provides them.

%========================================================
\section{Phrase-Conditioned Support Mask Extraction}
\label{app:mask-extraction}

This appendix provides the full specification of the mask extraction pipeline summarized in Section~\ref{sec:method-extraction}. 
The scoring criteria follow~\cite{kang2025your}, while the per-phrase application and soft aggregation are introduced in this work. 

For each training pair $(\mathcal{I}, c)$, we extract noun phrases $\{p_n\}_{n=1}^{N}$ from $c$ using spaCy, and collect text-to-image attention from every layer-head pair of the understanding path. 
We denote the resulting candidate map for phrase $p$ at layer $\ell$ and head $h$ as $A_{p,\ell,h} \in \mathbb{R}^{M}$, where $M$ is the number of image tokens. 
Candidate scoring and selection are performed independently for each phrase, since different phrases in the same image are typically grounded by different heads.

%--------------------------------------------------------
\subsection{Attention Map Preprocessing}
\label{app:preprocessing}

Each candidate map is preprocessed in two steps before scoring. 
Both steps are adopted from prior work and used as filtering only.

\paragraph{Mean-centered rectification}
We suppress weak background activations by subtracting the per-map mean and applying a non-negative rectification:
\begin{equation}
\tilde{A}_{p,\ell,h}(i) = \mathrm{ReLU}\!\left( A_{p,\ell,h}(i) - \frac{1}{M}\sum_{i'=1}^{M} A_{p,\ell,h}(i') \right).
\end{equation}
This step retains only above-average activations within each candidate map.

\paragraph{Visual sink removal}
Following \cite{kang2025see}, we account for visual attention sinks: a small set of image tokens that consistently absorb large attention mass irrespective of semantic content. 
We compute a global sink score $g(i)$ for each image token $i$ by averaging $\tilde{A}_{p,\ell,h}(i)$ across the entire candidate pool, and define the sink set $\mathcal{S}$ as the top $q$ fraction of tokens with the largest $g(i)$. 
For each candidate, we measure the non-sink mass ratio
\begin{equation}
r_{p,\ell,h} = \frac{\sum_{i \notin \mathcal{S}} \tilde{A}_{p,\ell,h}(i)}{\sum_{i=1}^{M} \tilde{A}_{p,\ell,h}(i) + \varepsilon}.
\end{equation}
Maps with $r_{p,\ell,h}$ below a threshold $\eta$ are discarded as sink-dominated. 
The remaining maps are sink-zeroed: % and renormalized:
\begin{equation}
\bar{A}_{p,\ell,h}(i) =
\begin{cases}
0 & i \in \mathcal{S}, \\[2pt]
\tilde{A}_{p,\ell,h}(i) & i \notin \mathcal{S}.
\end{cases}
\end{equation}
The preprocessed maps $\{\bar{A}_{p,\ell,h}\}$ form the surviving candidate pool $\mathcal{C}_p$ for each phrase $p$.

%--------------------------------------------------------
\subsection{Per-Phrase Scoring}
\label{app:scoring}

After preprocessing, each surviving candidate is scored on two criteria: relevance and spatial plausibility. 
Both scores are computed per phrase and standardized within the corresponding candidate pool $\mathcal{C}_p$, so that scores remain comparable across phrases.

\paragraph{Attention-sum score}
The attention-sum measures how much mass the candidate places on image tokens for the given phrase. 
We compute the raw sum
\begin{equation}
u_{p,\ell,h} = \sum_{i=1}^{M} \bar{A}_{p,\ell,h}(i),
\end{equation}
and standardize within $\mathcal{C}_p$:
\begin{equation}
s^{\mathrm{sum}}_{p,\ell,h} = \frac{u_{p,\ell,h} - \mu^{\mathrm{sum}}_p}{\sigma^{\mathrm{sum}}_p + \varepsilon},
\end{equation}
where $\mu^{\mathrm{sum}}_p$ and $\sigma^{\mathrm{sum}}_p$ are the mean and standard deviation of $\{u_{p, \ell', h'}\}_{(\ell', h') \in \mathcal{C}_p}$.

\paragraph{Entropy score}
The entropy term penalizes overly diffuse maps. 
We compute the spatial entropy of the map normalized to unit mass, so that the score depends on spread and not on mass:
\begin{equation}
    P_{p,\ell,h}(i) = \frac{\bar{A}_{p,\ell,h}(i)}{u_{p,\ell,h} + \varepsilon}, \qquad
H_{p,\ell,h} = -\sum_{i=1}^{M} P_{p,\ell,h}(i) \log\big(P_{p,\ell,h}(i) + \varepsilon\big)
\end{equation}
and standardize the negative entropy within the same candidate pool:
\begin{equation}
s^{\mathrm{ent}}_{p,\ell,h} = \frac{-H_{p,\ell,h} - \mu^{\mathrm{ent}}_p}{\sigma^{\mathrm{ent}}_p + \varepsilon},
\end{equation}
with $\mu^{\mathrm{ent}}_p$ and $\sigma^{\mathrm{ent}}_p$ defined analogously over $\{-H_{p, \ell', h'}\}$. 
This formulation favors candidates that are concentrated relative to other candidates for the same phrase, without imposing an absolute compactness assumption.

\paragraph{Final score}
We sum the two standardized terms with equal weighting:
\begin{equation}
s_{p,\ell,h} = s^{\mathrm{sum}}_{p,\ell,h} + s^{\mathrm{ent}}_{p,\ell,h}.
\end{equation}
Once candidate scoring is performed per phrase, the two standardized terms operate on comparable scales and play complementary roles, with the sum term capturing relevance and the entropy term penalizing diffuse maps. 

%--------------------------------------------------------
\subsection{Soft Top-$K$ Aggregation}
\label{app:aggregation}

For each phrase $p$, we rank candidates by $s_{p,\ell,h}$ and retain the top $K$ maps $\{\bar{A}_{p,k}\}_{k=1}^{K}$ with corresponding scores $\{s_{p,k}\}_{k=1}^{K}$. The phrase-conditioned support mask is obtained by temperature-controlled softmax aggregation:
\begin{equation}
A_p = \sum_{k=1}^{K} w_{p,k}\,\bar{A}_{p,k}, \qquad
w_{p,k} = \frac{\exp(s_{p,k} / \tau)}{\sum_{k'=1}^{K} \exp(s_{p,k'} / \tau)}.
\end{equation}
The aggregated mask is renormalized so that $\sum_{i=1}^{M} A_p(i) = 1$. 
Compared to hard top-$K$ averaging, soft aggregation preserves informative sub-maximal candidates and produces a denser support, which is particularly useful in the per-phrase regime where the set of useful heads varies across phrases.

%--------------------------------------------------------
\subsection{Hyperparameter Values}
\label{app:extraction-hparams}
\begin{table}[t]
\centering
\caption{
    Hyperparameters for mask extraction.
}
\vspace{0.2cm}
\scalebox{0.8}{
\begin{tabular}{l c l}
\toprule
Hyperparameter & Symbol & Value \\
\midrule
Sink fraction          & $q$       & 0.05 \\
Non-sink mass threshold & $\eta$       & 0.7 \\
Top-$K$ candidates     & $K$          & 8 \\
Aggregation temperature & $\tau$       & 0.07 \\
Numerical stabilizer   & $\varepsilon$ & $10^{-8}$ \\
\bottomrule
\end{tabular}
}
\end{table}
These values are fixed across all backbones in our experiments. 
Per-phrase candidate selection allows the pipeline to use a single set of hyperparameters without backbone-specific tuning.

%--------------------------------------------------------
\subsection{Output Interpretation}
\label{app:interpretation}

The extracted $A_p$ should be understood as a phrase-conditioned soft spatial support, not as a hard segmentation or an exact instance decomposition. The masks are reliable at coarse localization and at separating spatial responsibility across phrases, and less suitable for recovering finer structural properties such as exact cardinality. We treat $A_p$ as a frozen teacher signal derived from the understanding path, and use it to supervise the generation path through phrase-local contrastive alignment (Section~\ref{sec:method-contrastive}) and structural scaffolding (Section~\ref{sec:method-scaffold}).

%========================================================
\section{Dataset and Evaluation Protocol}
\label{app:data-eval}

\subsection{Training Dataset}
\label{app:train-data}
% Echo-4o Instruction-Following subset
We use the Instruction-Following subset of Echo-4o~\cite{ye2025echo}, a synthetic dataset generated by GPT-4o that contains approximately 70K image-caption pairs. 
The captions are constructed through template-driven composition over curated base concepts, covering compositional instructions involving multiple objects, attributes (color, material, pattern), counts, sizes, and spatial relations. 
The synthetic and template-controlled nature of Echo-4o provides clean, diverse compositional supervision that complements the underrepresentation of dense compositional prompts in real-world image-caption corpora. 
We use the dataset under its publicly released license and apply no additional preprocessing beyond noun phrase extraction with spaCy as described in Appendix~\ref{app:mask-extraction}.

\subsection{GenEval2 Protocol}
\label{app:geneval2}
% Atomic question categories with examples
% Oracle MLLM judge (Qwen3-VL-8B): prompt template, generation settings
% Soft-TIFA scoring formula
% Geometric mean computation, atomicity binning
GenEval2~\cite{kamath2025geneval} contains 800 prompts stratified by atomicity from 3 to 10, with 100 prompts per level. 
Atomicity counts the number of visual atoms (objects, attributes, relations) specified in a prompt. 
Each prompt is decomposed into atomic VQA questions covering five skill categories (Object, Attribute, Count, Position, Verb), and generated images are evaluated by Soft-TIFA, an atom-level VQA judge based on Qwen3-VL-8B~\cite{bai2025qwen3}.
We follow the standard protocol and generate one image per prompt with default sampling hyperparameters of each backbone. 
We report arithmetic mean (AM), geometric mean (GM), category-level scores, and atomicity-stratified GM as defined in the original benchmark.

\subsection{Oracle Image Set Curation for Section~\ref{sec:motivation}}
\label{app:oracle}

The understanding-path diagnostic in Section~\ref{sec:motivation} requires images that satisfy every atomic condition of a given prompt. 
GenEval2~\cite{kamath2025geneval} provides prompts and atomic questions but does not release paired images, so we construct an oracle image set ourselves. 
The set is used solely as input to the understanding path during the diagnostic in Section~\ref{sec:motivation}, and is not used to train any of the compared models or to evaluate generation quality.

\paragraph{Prompts}
We use the GenEval2 prompts unchanged. 
The set spans atomicity levels from 3 to 10 with 100 prompts at each level, for a total of 800 prompts. 
Each prompt is paired with a single oracle image, yielding a final oracle set of 800 image-prompt pairs.

\paragraph{Candidate generation}
For each prompt, we generate candidate images using two strong image generation models, GPT-Image-1 and Nano Banana. 
Using two independent generators reduces bias toward the stylistic and compositional preferences of a single model, and increases the likelihood that at least one candidate is close to satisfying the prompt before any editing is applied.

\paragraph{Atomic condition verification}
Each candidate is evaluated by the same oracle MLLM judge used in the GenEval2 protocol, Qwen3-VL-8B~\cite{bai2025qwen3}, against all atomic questions of the corresponding prompt. 
Candidates that satisfy every atomic condition are retained as oracle images. 
Candidates that fail one or more conditions enter the editing stage.

\paragraph{Model-based editing}
We edit partially satisfying candidates using the same two models (GPT-Image-1 and Nano Banana) in their image-editing modes. 
Edits are conditioned on a description of the failing atomic condition. 
After each edit, the result is re-evaluated by the MLLM judge, and the editing loop terminates when all atomic conditions are satisfied or after a small number of unsuccessful iterations. 
Images that cannot be brought to full satisfaction are discarded rather than retained with partial coverage. 
No human editing is performed at any stage.

\paragraph{Visual inspection by the authors}
The authors visually inspect every retained image in the final oracle set. 
The role of inspection is to catch failure modes that the MLLM judge may miss, such as subtle attribute leakage, unnatural artifacts introduced by editing, or duplicated objects that nominally satisfy a count condition. 
Images that pass MLLM verification but fail authorial inspection are returned to the editing stage or replaced with a freshly generated candidate. 
No external workers are involved in either editing or inspection.

\begin{figure}[t]
\centering
\setlength{\tabcolsep}{2pt}
\renewcommand{\arraystretch}{1.0}
\begin{tabular}{@{}
>{\centering\arraybackslash}m{0.22\textwidth}
>{\centering\arraybackslash}m{0.22\textwidth}
>{\centering\arraybackslash}m{0.22\textwidth}
>{\centering\arraybackslash}m{0.22\textwidth}
@{}}
\sffamily\footnotesize Atomicity 3
& \sffamily\footnotesize Atomicity 5
& \sffamily\footnotesize Atomicity 7
& \sffamily\footnotesize Atomicity 9
\\
\includegraphics[width=0.21\textwidth]{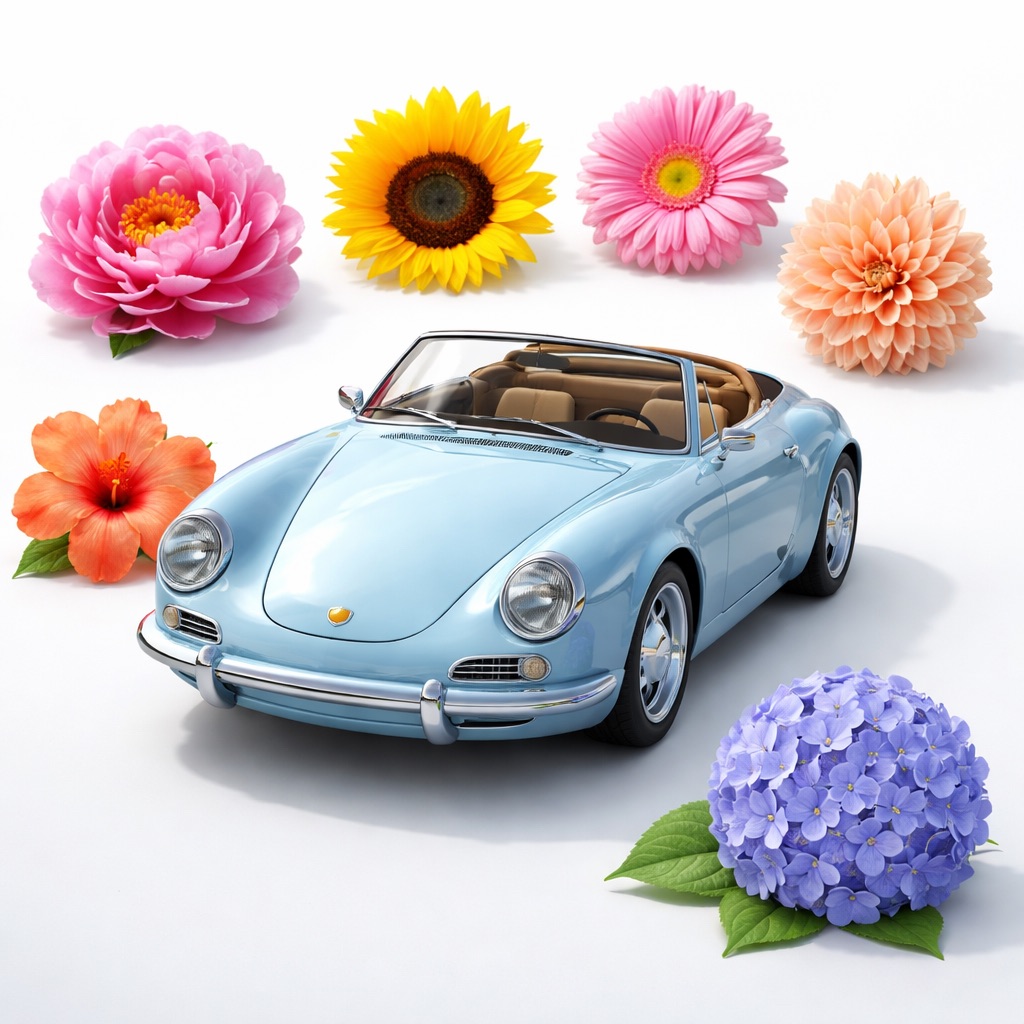}
& \includegraphics[width=0.21\textwidth]{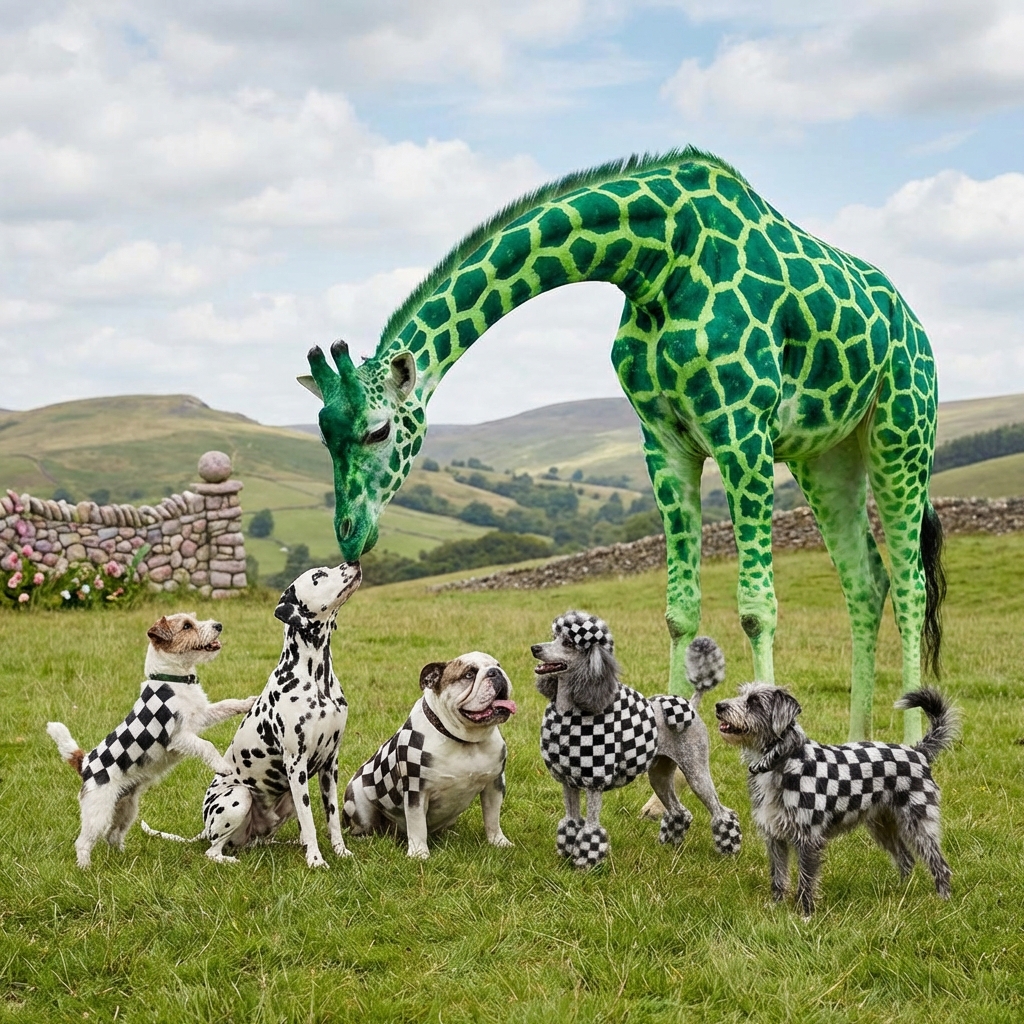}
& \includegraphics[width=0.21\textwidth]{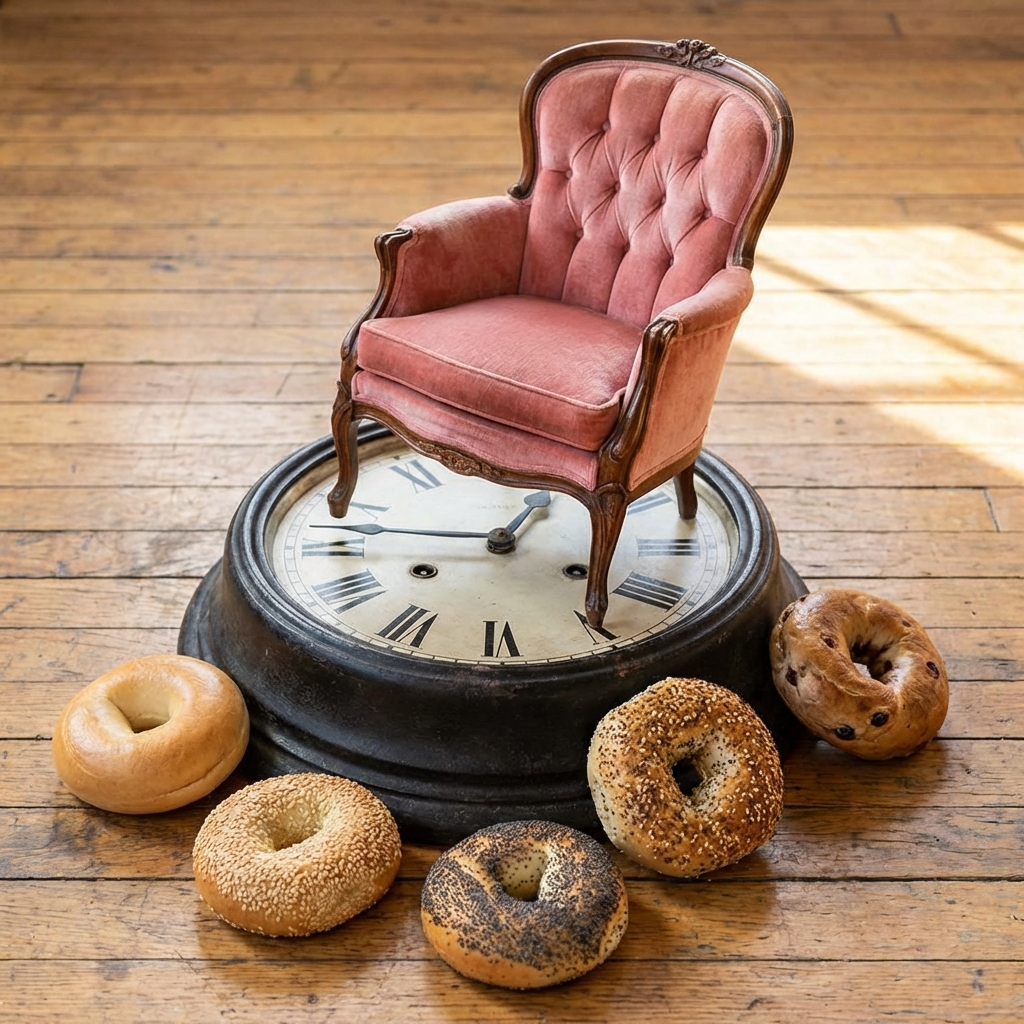}
& \includegraphics[width=0.21\textwidth]{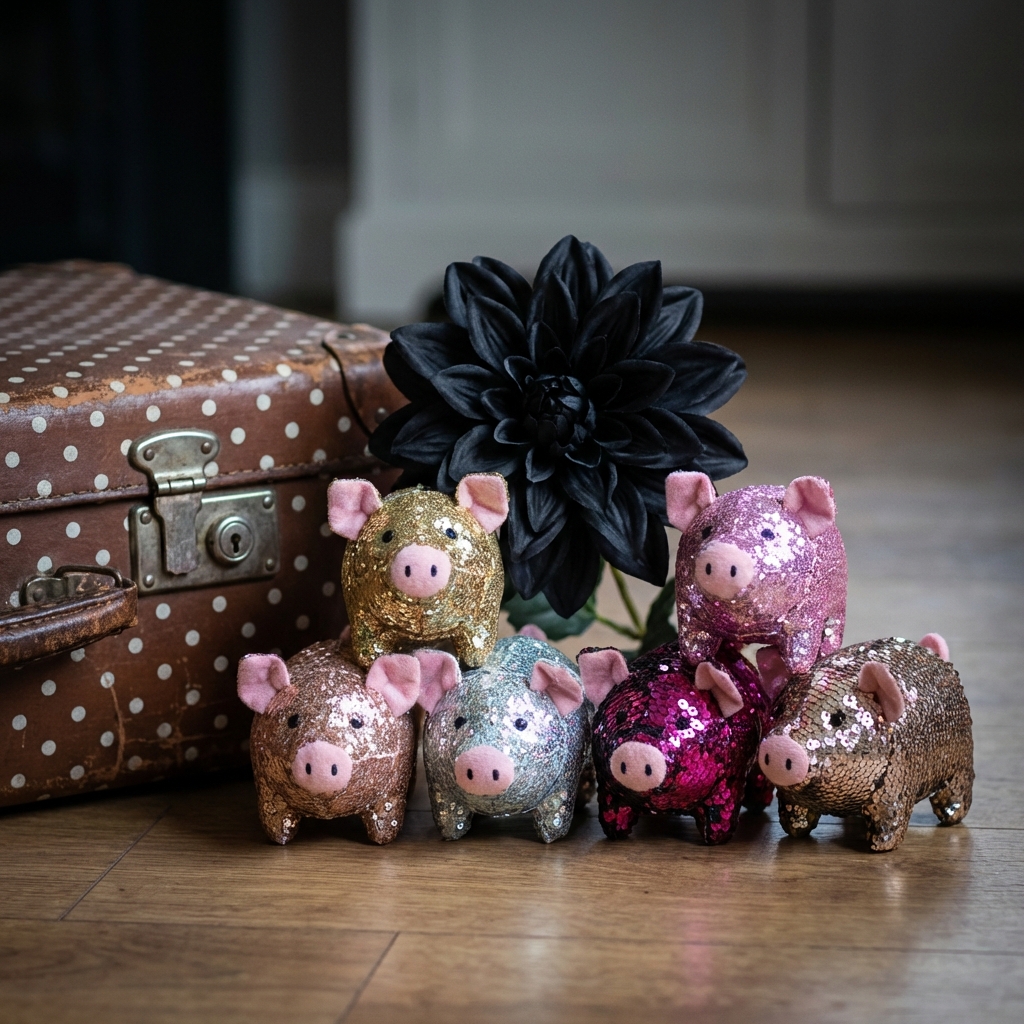}
\\
\sffamily\scriptsize ``a car and six flowers''
& \sffamily\scriptsize ``five checkered dogs and a green giraffe''
& \sffamily\scriptsize ``a pink chair on top of a black clock, and five bagels''
& \sffamily\scriptsize ``six sparkling pigs in front of a black flower to the right of a spotted suitcase"
\\
\end{tabular}
\caption{
Representative oracle images across atomicity levels. 
Each image satisfies all atomic conditions of its paired GenEval2 prompt under both the MLLM judge and the authors' visual inspection.
}
\label{fig:oracle_examples}
\end{figure}

\paragraph{Final set statistics}
The final oracle set contains 800 images, distributed evenly across atomicity levels with 100 images per level. 
Each retained image satisfies all atomic conditions of its paired prompt under both the MLLM judge and the authors' visual inspection.
Figure~\ref{fig:oracle_examples} shows representative examples across atomicity levels.

\paragraph{Scope and limitations.}
The oracle set is a diagnostic instrument rather than a benchmark. 
It is constructed at a scale appropriate for probing understanding-path behavior across atomicity levels, not for training or for benchmark-grade evaluation. 
Both the MLLM judge and the editing models are fallible, and a small number of retained images may exhibit subtle inconsistencies that escape both verification stages. 
We accept this as the same noise floor that affects the GenEval2 generation evaluation, and the diagnostic in Section~\ref{sec:motivation} should be interpreted in this context. 
The oracle set is not used as training data or as a generation benchmark anywhere in the paper.

\begin{table}[t]
\centering
\caption{
Performance comparison on GenEval++~\cite{ye2025echo}. 
We report per-task accuracy (\%) and overall accuracy. 
Task abbreviations: CA (color attribute), SCA (spatial-count attribute), CSA (color-spatial attribute), CC (color-count), MC (multi-count), SS (size/spatial), Ct (counting). 
Bold indicates the best result per backbone group.
}
\vspace{0.2cm}
\label{tab:genevalpp}
\setlength{\tabcolsep}{4pt}
\small
\scalebox{0.8}{
\begin{tabular}{l c c c c c c c c}
\toprule
Method & CA & SCA & CSA & CC & MC & SS & Ct & Overall \\
\midrule
\multicolumn{9}{l}{\textit{BLIP-3o 4B}} \\
Base       & 12.50 & 17.50 & 40.00 & 15.00 & 15.00 & 52.50 & 27.50 & 25.71 \\
+ LoRA FT  & 12.50 & 20.00 & \textbf{42.50} & 15.00 & 12.50 & 55.00 & 27.50 & 26.43 \\
+ SST (Ours) & \textbf{20.00} & \textbf{25.00} & 40.00 & \textbf{17.50} & \textbf{17.50} & \textbf{62.50} & \textbf{32.50} & \textbf{30.71} \\
\midrule
\multicolumn{9}{l}{\textit{BLIP-3o 8B}} \\
Base       & 15.00 & \textbf{22.50} & 45.00 & 10.00 & 25.00 & 62.50 & \textbf{35.00} & 30.71 \\
+ LoRA FT  & 17.50 & 20.00 & 47.50 & 12.50 & 30.00 & 60.00 & \textbf{35.00} & 31.79 \\
+ SST (Ours) & \textbf{17.50} & \textbf{22.50} & \textbf{50.00} & \textbf{15.00} & \textbf{30.00} & \textbf{67.50} & 32.50 & \textbf{33.57} \\
\midrule
\multicolumn{9}{l}{\textit{Show-o2 7B}} \\
Base       & 27.50 & \textbf{7.50} & 17.50 & 12.50 & 12.50 & 30.00 & 10.00 & 16.79 \\
+ LoRA FT  & 30.00 & 5.00 & 17.50 & 15.00 & 17.50 & 30.00 & 12.50 & 18.21 \\
+ SST (Ours) & \textbf{32.50} & 5.00 & \textbf{20.00} & \textbf{15.00} & \textbf{17.50} & \textbf{35.00} & \textbf{12.50} & \textbf{19.64} \\
\midrule
\multicolumn{9}{l}{\textit{BAGEL}} \\
Base       & 45.00 & 7.50 & 37.50 & 27.50 & \textbf{52.50} & 50.00 & 47.50 & 38.21 \\
+ LoRA FT  & 55.00 & 17.50 & \textbf{50.00} & 37.50 & 42.50 & \textbf{55.00} & \textbf{57.50} & 45.00 \\
+ SST (Ours) & \textbf{60.00} & \textbf{20.00} & \textbf{50.00} & \textbf{40.00} & 47.50 & \textbf{55.00} & 55.00 & \textbf{46.79} \\
\bottomrule
\end{tabular}
}
\end{table}

%========================================================
\section{Additional Quantitative Results}
\label{app:additional-quant}

\subsection{GenEval++}
\label{app:other-benchmarks}
GenEval++~\cite{ye2025echo} contains 280 prompts spanning seven task types of increasing difficulty, evaluated by a GPT-4.1-based judge that marks an image correct only if all compositional constraints are satisfied. 
Table~\ref{tab:genevalpp} reports per-task and overall accuracy across four UMM backbones. 
%Across all backbones, our framework consistently improves over both the base UMM and LoRA fine-tuning. 
SST improves overall accuracy over both the base UMM and LoRA fine-tuning on all four backbones.
The ranking Base $<$ LoRA FT $<$ SST holds across all backbones, indicating that the improvements observed on GenEval2 transfer consistently to this independent compositional benchmark.

\begin{table}[t]
\centering
\caption{
Mask quality ablation on a 1K Echo-4o subset (BLIP-3o 4B). 
We progressively add per-phrase head selection and soft top-$K$ aggregation to the global head selection baseline of~\cite{kang2025your}. 
Sink suppression~\cite{kang2025see} is applied to all variants. 
Bold indicates the best result in each column.
}
\vspace{0.2cm}
\label{tab:mask_quality}
\scalebox{0.8}{
\begin{tabular}{c c c c}
\toprule
Per-phrase selection & Soft scoring & AIR & Pt \\
\midrule
                     &              & 0.321 & 0.603 \\
\checkmark           &              & 0.371 & 0.645 \\
\checkmark           & \checkmark   & \textbf{0.411} & \textbf{0.718} \\
\bottomrule
\end{tabular}
}
\end{table}

\begin{table}[t]
\centering
\caption{Sensitivity to the extraction hyperparameters on BLIP-3o 4B (single run each). 
AIR and Pt are measured against Grounded-SAM pseudo-references on the 1K Echo-4o subset of Table~\ref{tab:mask_quality}. 
AM and GM are GenEval2 scores after retraining with the resulting supports. 
Our setting is marked in gray.}
\vspace{0.2cm}
\label{tab:extraction_sweep}
\scalebox{0.8}{
\begin{tabular}{lcccc}
\toprule
 & AIR & Pt & AM & GM \\
\midrule
\multicolumn{5}{l}{\textit{Temperature $\tau$ ($K = 8$)}} \\
$\tau = 0.01$ & 0.341 & 0.643 & 56.28 & 14.87 \\
\rowcolor{gray!15} $\tau = 0.07$ & 0.411 & 0.718 & 58.34 & 16.05 \\
$\tau = 1$    & 0.433 & 0.726 & 58.26 & 16.01 \\
\midrule
\multicolumn{5}{l}{\textit{Top-$K$ ($\tau = 0.07$)}} \\
$K = 1$  & 0.373 & 0.661 & 55.58 & 14.37 \\
$K = 4$  & 0.403 & 0.693 & 57.62 & 15.82 \\
\rowcolor{gray!15} $K = 8$  & 0.411 & 0.718 & 58.34 & 16.05 \\
$K = 16$ & 0.432 & 0.732 & 58.38 & 16.13 \\
\bottomrule
\end{tabular}}
\end{table}

\subsection{Ablation Studies for Support Mask}
\label{app:mask-quality}
% Comparison with Grounded-SAM pseudo-reference
% Pointing accuracy, activation-inside ratio
% Ablation: raw attention vs global head vs full pipeline

\paragraph{Support Mask Quality.}
We verify mask quality on a uniformly sampled 1{,}000 subset of Echo-4o, using Grounded-SAM~\cite{ren2024grounded} on the same noun phrases as pseudo-references. 
We report activation-inside ratio (AIR) and center-of-mass pointing (Pt), which align with the use case of our supports as soft pooling regions, and focus on relative improvements since the pseudo-references are themselves imperfect. 
As shown in Table~\ref{tab:mask_quality}, both per-phrase head selection and soft top-$K$ aggregation provide consistent improvements in AIR and pointing accuracy over global head selection~\cite{kang2025your}.

\paragraph{Sensitivity to extraction hyperparameters.}
We retrained BLIP-3o 4B with supports extracted under different values of $\tau$ and $K$ (Table~\ref{tab:extraction_sweep}). 
Aggregation must draw on more than one candidate map. 
Setting $K = 1$, or $\tau = 0.01$, which concentrates the softmax on the top-ranked map, lowers GM by more than one point. 
For $\tau \geq 0.07$ and $K \geq 8$, GM varies by at most 0.12. 
AIR and pointing accuracy keep improving in this range, but generation scores do not.

\begin{table}[t]
\centering
\caption{Prompt-level paired bootstrap of SST over LoRA FT on GenEval2 (800 prompts, $B = 10{,}000$, percentile intervals), for one training run per backbone. 
Win rate is the fraction of prompts on which SST scores higher.}
\label{tab:bootstrap}
\vspace{0.2cm}
\scalebox{0.8}{
\begin{tabular}{llccc}
\toprule
Backbone & Metric & $\Delta$ (SST $-$ LoRA FT) & 95\% CI & Win rate \\
\midrule
\multirow{2}{*}{BLIP-3o 4B} & AM & $+3.78$ & [$+2.69$, $+4.86$] & 59.4\% \\
                            & GM & $+2.70$ & [$+1.11$, $+4.28$] & 60.9\% \\
\midrule
\multirow{2}{*}{BAGEL}      & AM & $+1.04$ & [$+0.27$, $+1.81$] & 56.4\% \\
                            & GM & $+0.82$ & [$+0.11$, $+1.53$] & 53.9\% \\
\bottomrule
\end{tabular}}
\end{table}

\begin{table}[t]
\centering
\caption{Human evaluation of LoRA FT and SST on 80 GenEval2 prompts per backbone, 10 per atomicity level, sampled with a fixed seed before any scores were inspected.
%sampled uniformly within each atomicity level with a fixed seed before inspecting any scores. 
Human AM is the mean majority-voted satisfaction rate over atom questions.
All-atom rate is the fraction of prompts on which the human majority marks every atom question as satisfied.
Soft-TIFA scores are computed on the same subset and therefore differ from Table~\ref{tab:geneval2-main}. 
On BAGEL, this subset shows a larger Soft-TIFA gap (+9.0 AM) than the full set (+1.04; Table~\ref{tab:bootstrap}).
Agreement is the fraction of atom questions on which Soft-TIFA matches the human majority, computed separately on the images of each variant. 
Brackets are 95\% prompt-level paired bootstrap intervals. $\Delta$ is computed before rounding.}
\label{tab:human_eval}
\vspace{0.2cm}
\scalebox{0.8}{
\begin{tabular}{llccl}
\toprule
Backbone & Metric & LoRA FT & SST & $\Delta$ [95\% CI] \\
\midrule
\multirow{5}{*}{BLIP-3o 4B}
 & AM (human)            & 58.3 & 62.6 & $+4.4$ [$+1.5$, $+7.1$] \\
 & AM (Soft-TIFA)        & 56.1 & 59.4 & $+3.3$ \\
 \cmidrule(l){2-5}
 & All-atom rate (human) & 6.3  & 8.8  & $+2.5$ [$-1.3$, $+5.8$] \\
 & GM (Soft-TIFA)        & 9.5  & 11.5 & $+2.0$ \\
 \cmidrule(l){2-5}
 & Agreement             & 88.2 & 87.8 & $-0.5$ \\
\midrule
\multirow{5}{*}{BAGEL}
 & AM (human)            & 70.0 & 76.9 & $+7.0$ [$+2.7$, $+11.1$] \\
 & AM (Soft-TIFA)        & 66.4 & 75.4 & $+9.0$ \\
 \cmidrule(l){2-5}
 & All-atom rate (human) & 12.5 & 18.8 & $+6.3$ [$-1.3$, $+14.2$] \\
 & GM (Soft-TIFA)        & 20.2 & 28.8 & $+8.6$ \\
 \cmidrule(l){2-5}
 & Agreement             & 90.0 & 89.9 & $-0.2$ \\
\bottomrule
\end{tabular}}
\end{table}

\subsection{Paired Bootstrap over Prompts.}
\label{app:bootstrap}
We test whether the gains of SST over LoRA FT exceed prompt sampling variation. 
For one training run per backbone, we resample the 800 GenEval2 prompts with replacement (B = 10,000) and compute paired per-prompt differences. 
All four intervals exclude zero (Table~\ref{tab:bootstrap}). 
The win rate is the fraction of prompts on which SST scores higher; no prompt is tied.

\subsection{Human Evaluation}
\label{app:human_eval}
We validate Soft-TIFA and the improvement of SST with a blinded human evaluation on BLIP-3o 4B and BAGEL. 
For each backbone, we sample 80 GenEval2 prompts, 10 per atomicity level, with a fixed seed before inspecting any scores. 
Three non-author annotators judge each image of LoRA FT and SST against the unchanged GenEval2 atom questions. 
Each image is presented independently, with model identity hidden and presentation order randomized. 
We take the majority vote per atom.

Soft-TIFA agrees with the human majority on 89\% of the 2,448 atom questions (Cohen’s $\kappa$ = 0.76), on both LoRA FT and SST outputs. 
Human judgments favor SST at the atom level by 4.4 points on BLIP-3o 4B and 7.0 points on BAGEL, and both paired bootstrap intervals exclude zero (Table~\ref{tab:human_eval}).
The all-atom rate also increases on both backbones, but its intervals include zero at this sample size.

%========================================================
\section{Additional Qualitative Results}
\label{app:qualitative}

\subsection{Phrase-Conditioned Support Mask Visualizations}
\label{app:mask_viz}

\begin{figure}[t]
\centering
\setlength{\tabcolsep}{2pt}
\renewcommand{\arraystretch}{1.0}
\begin{tabular}{@{}
>{\centering\arraybackslash}m{0.22\textwidth}
>{\centering\arraybackslash}m{0.22\textwidth}
>{\centering\arraybackslash}m{0.22\textwidth}
>{\centering\arraybackslash}m{0.22\textwidth}
@{}}
& \sffamily\footnotesize ``A baseball glove''
& \sffamily\footnotesize ``Three tigers''
& \sffamily\footnotesize ``Three pizzas''
\\
\includegraphics[width=0.21\textwidth]{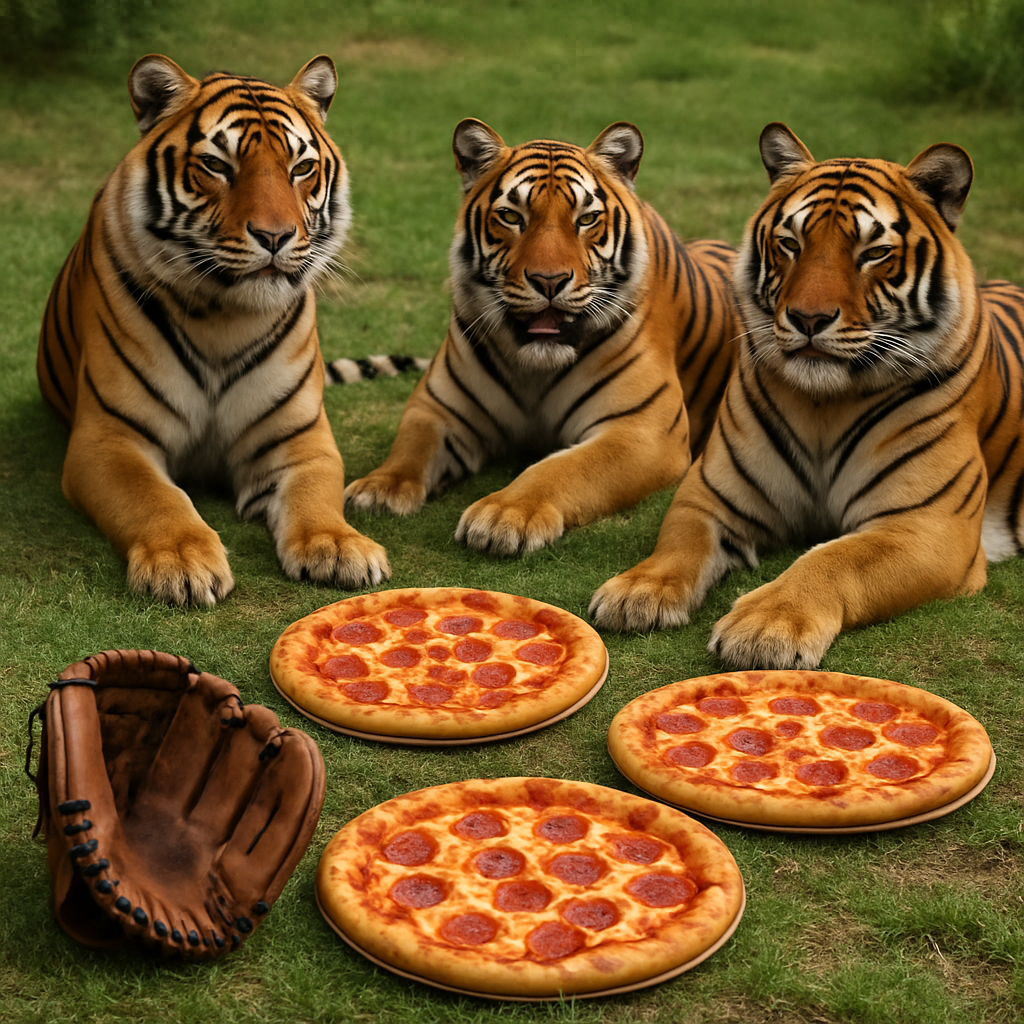}
& \includegraphics[width=0.21\textwidth]{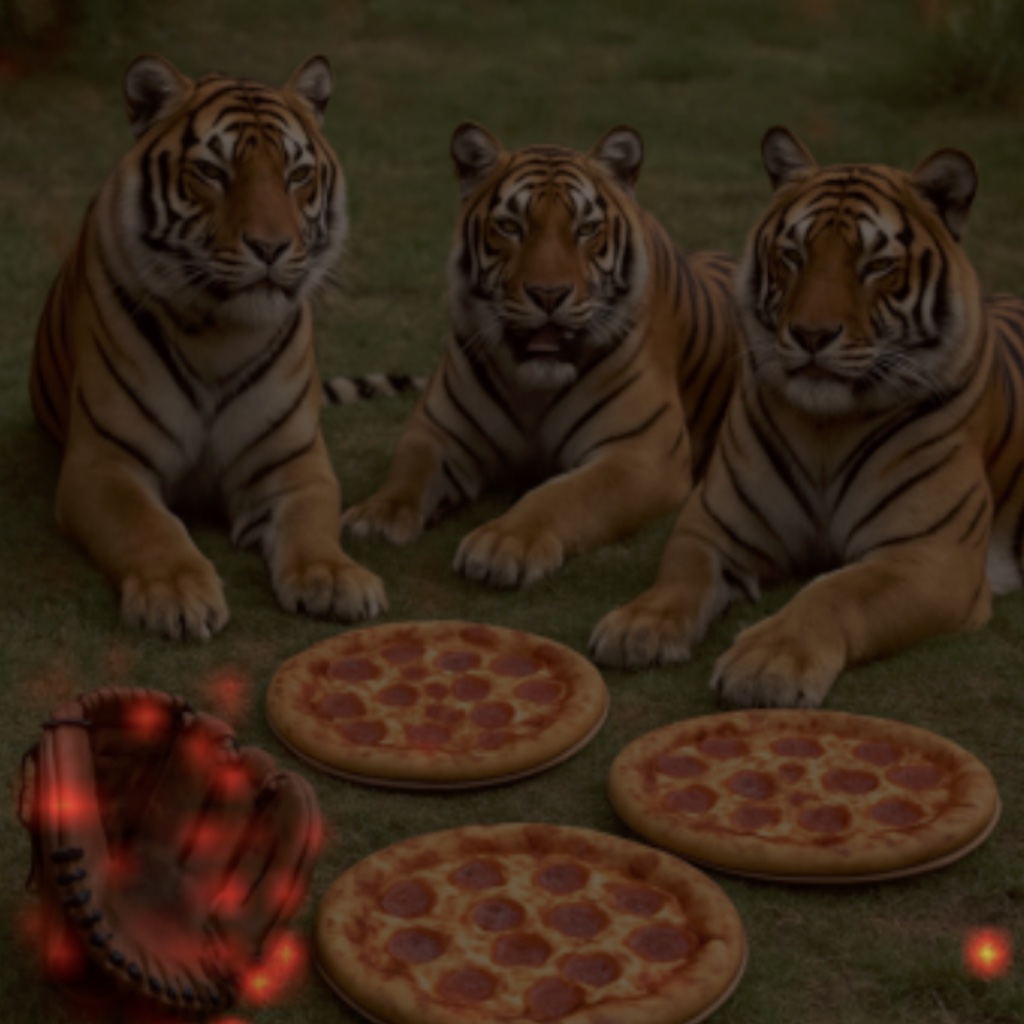}
& \includegraphics[width=0.21\textwidth]{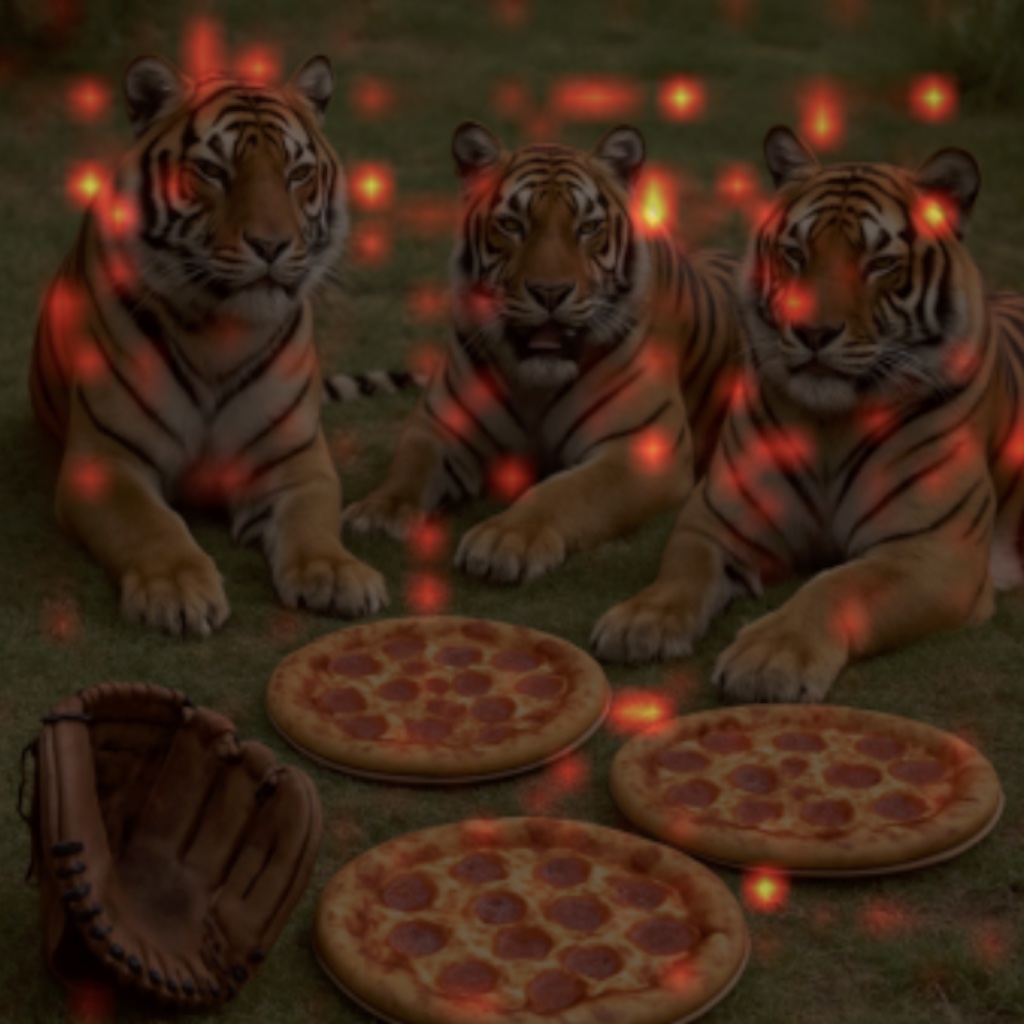}
& \includegraphics[width=0.21\textwidth]{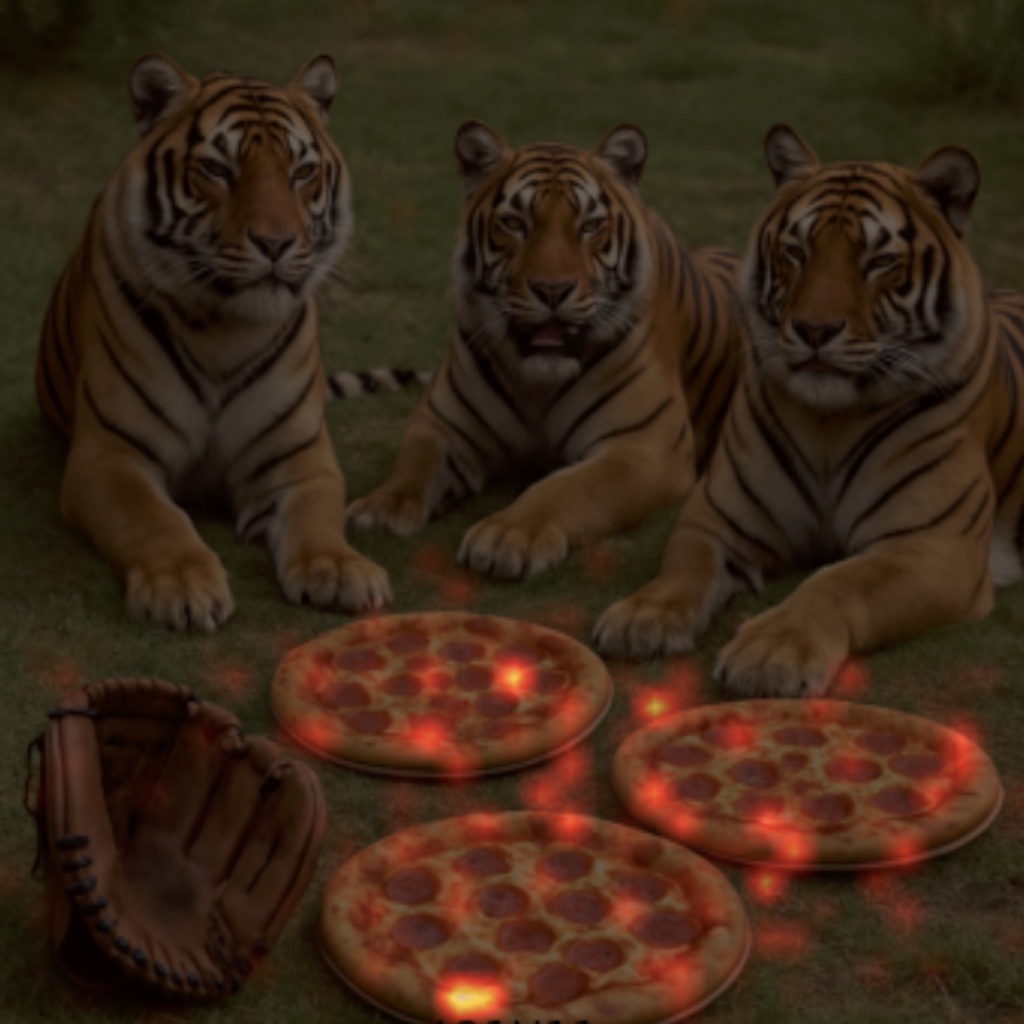}
\\[4pt]
& \sffamily\footnotesize ``A purple cow''
& \sffamily\footnotesize ``six dogs''
& \sffamily\footnotesize ``A croissant''
\\
\includegraphics[width=0.21\textwidth]{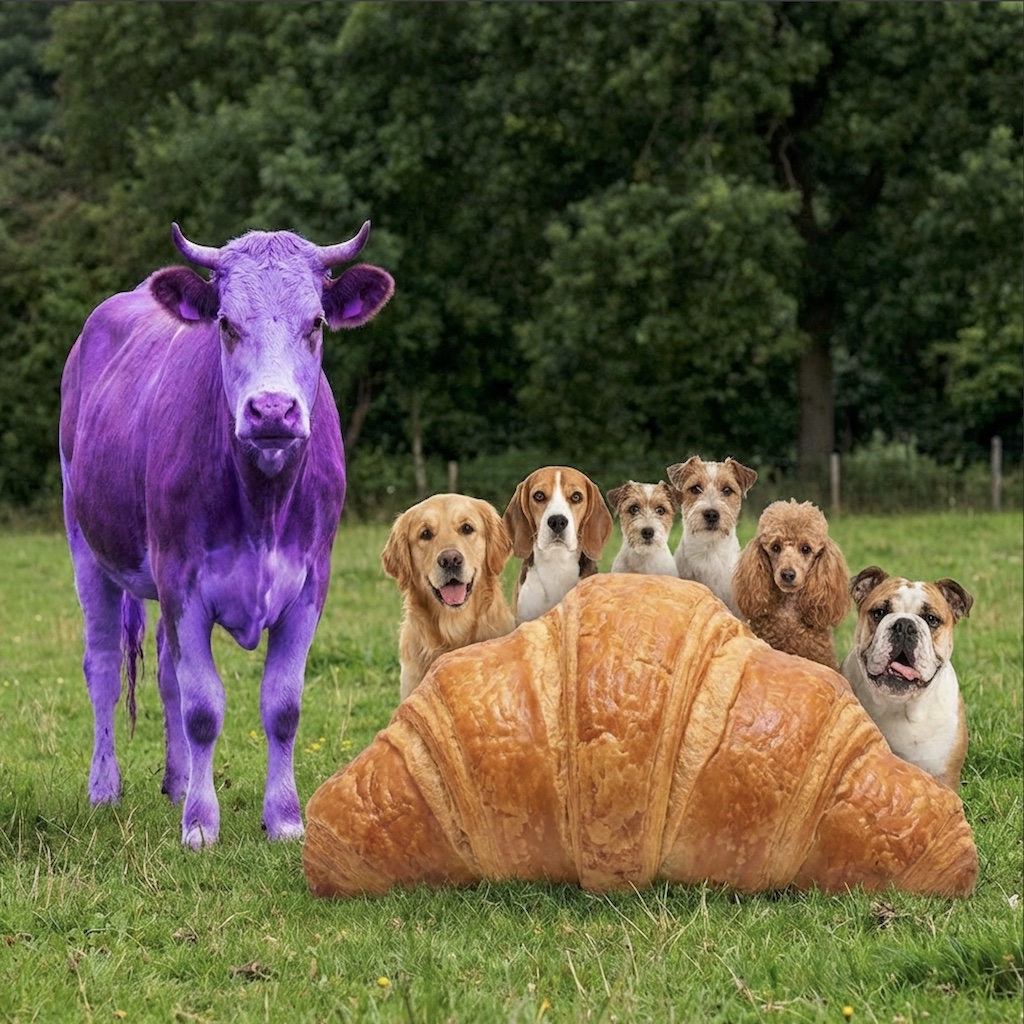}
& \includegraphics[width=0.21\textwidth]{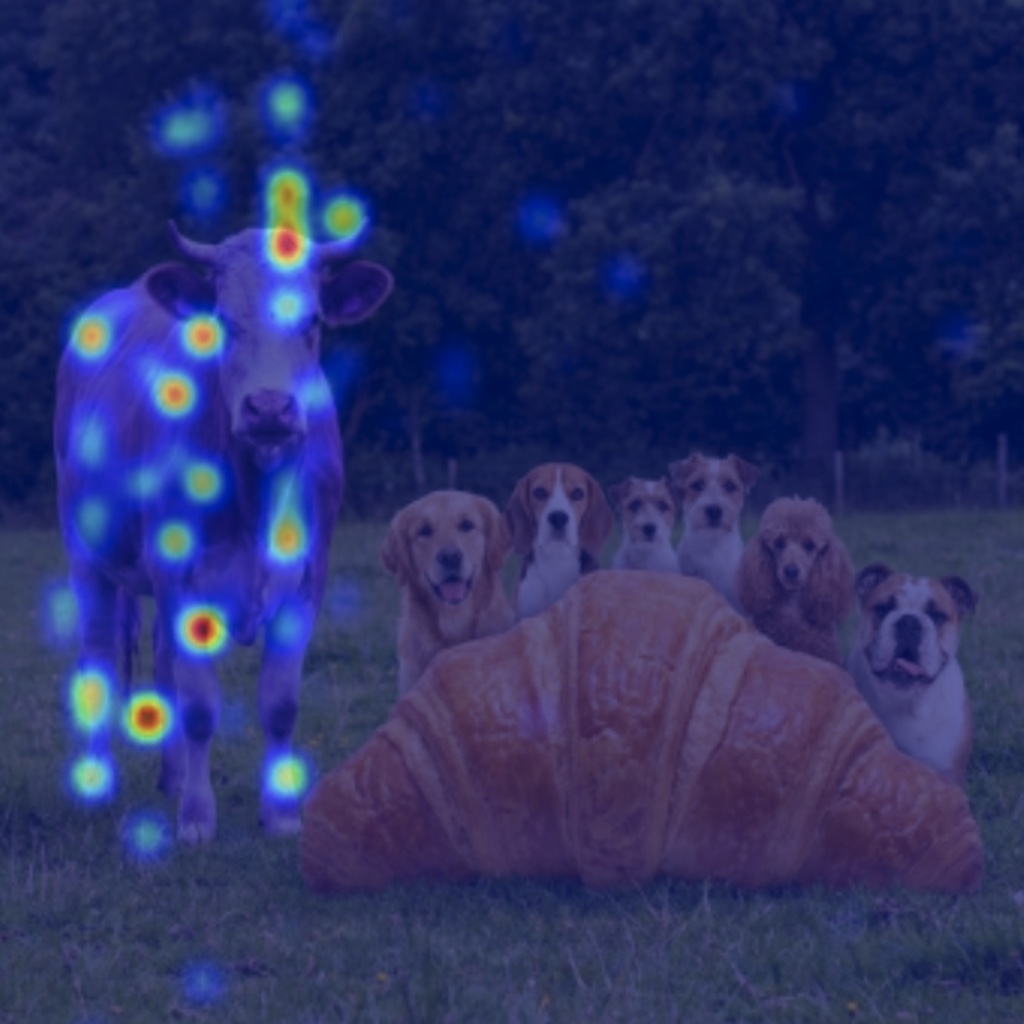}
& \includegraphics[width=0.21\textwidth]{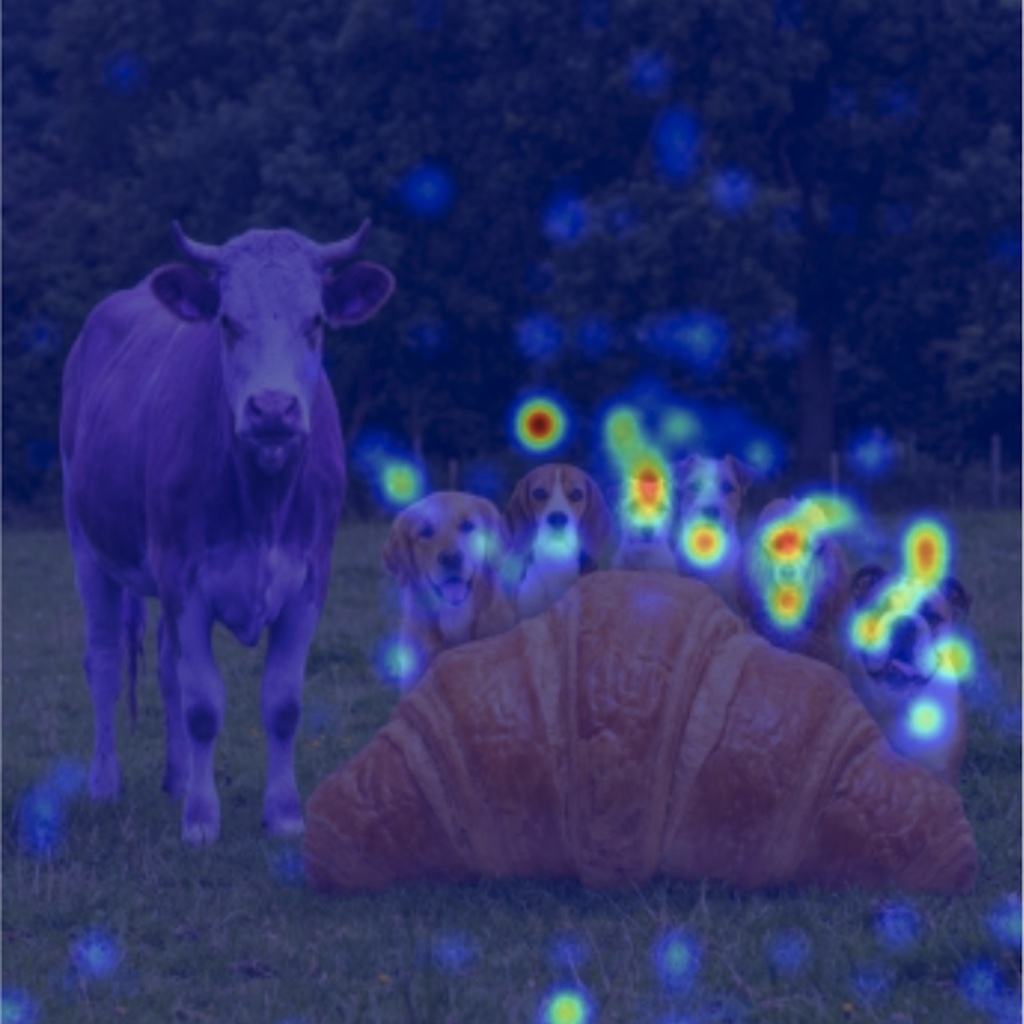}
& \includegraphics[width=0.21\textwidth]{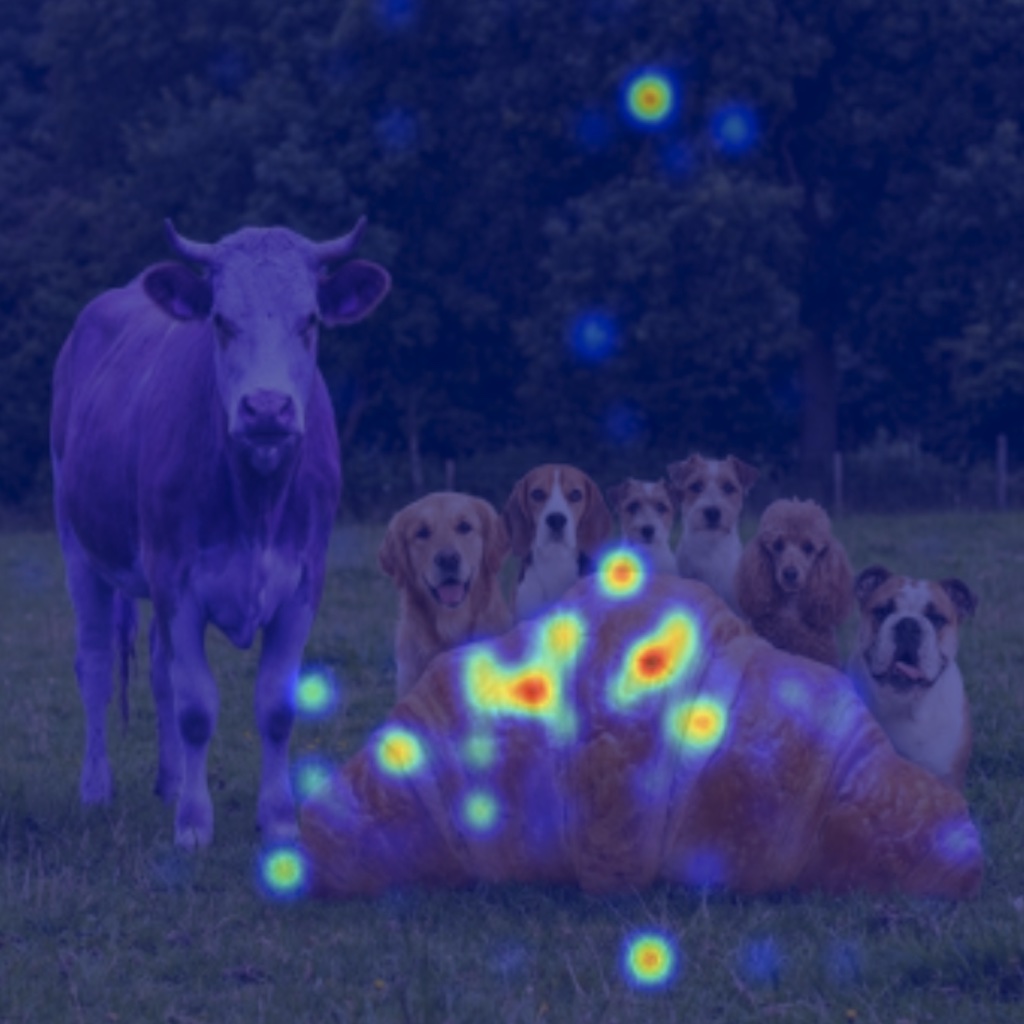}
\\
\end{tabular}
\caption{
Phrase-conditioned support masks on images from the training distribution (Echo-4o; top) and the evaluation distribution (GenEval2 oracle image; bottom). 
The masks form spatially coherent regions across both distributions, supporting their use as pooling regions for phrase-local alignment.
}
\label{fig:mask_viz}
\end{figure}

Figure~\ref{fig:mask_viz} illustrates the phrase-conditioned support masks extracted from the understanding path, complementing the schematic illustration in Figure~\ref{fig:teaser}. 
The masks are overlaid on the corresponding images, with brighter regions indicating stronger phrase support.

\subsection{Failure Cases}
\label{app:failure}

\begin{figure}[t]
\centering
\setlength{\tabcolsep}{2pt}
\renewcommand{\arraystretch}{1.0}
\begin{tabular}{@{}
>{\centering\arraybackslash}m{0.30\textwidth}
>{\centering\arraybackslash}m{0.30\textwidth}
>{\centering\arraybackslash}m{0.30\textwidth}
@{}}
\sffamily\footnotesize Base
& \sffamily\footnotesize LoRA FT
& \sffamily\footnotesize SST (Ours)
\\
\includegraphics[width=0.29\textwidth]{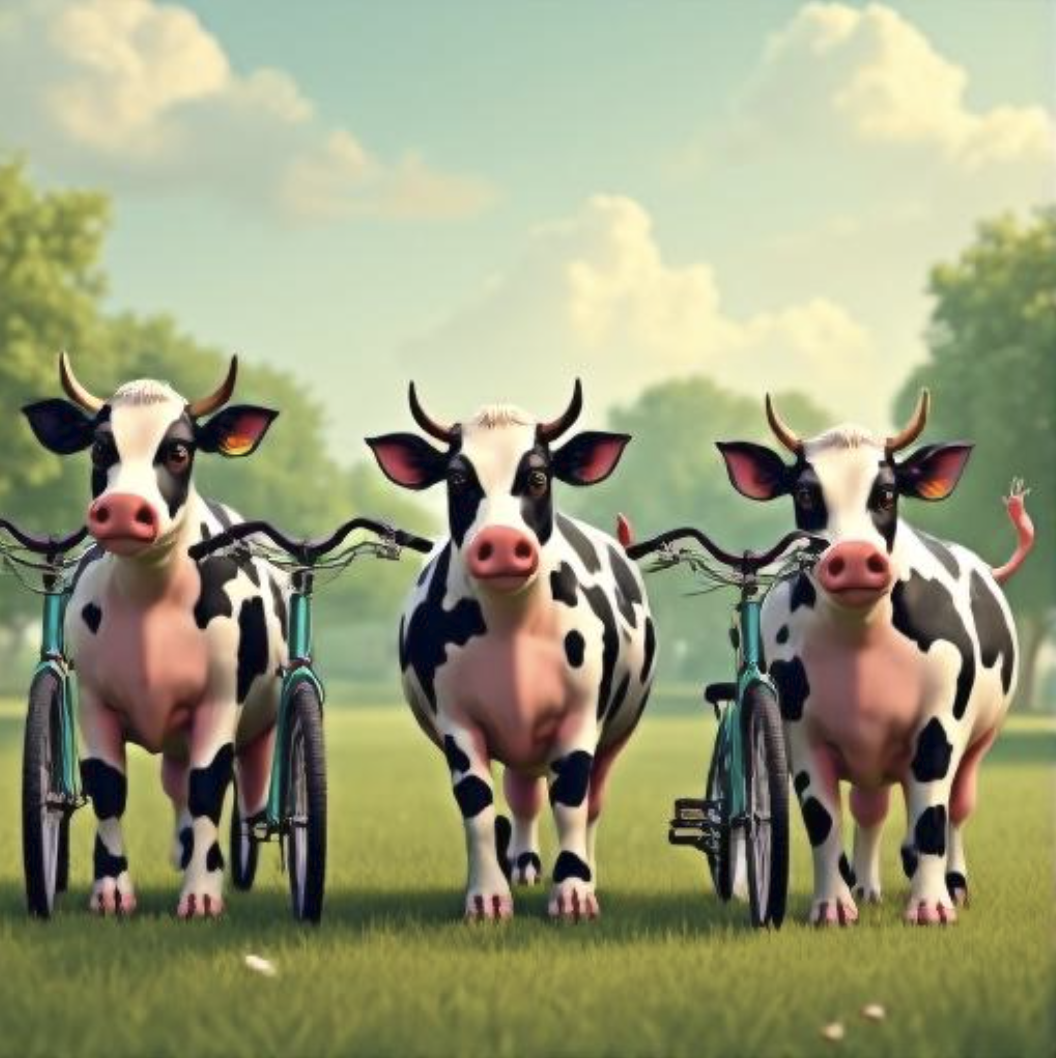}
& \includegraphics[width=0.29\textwidth]{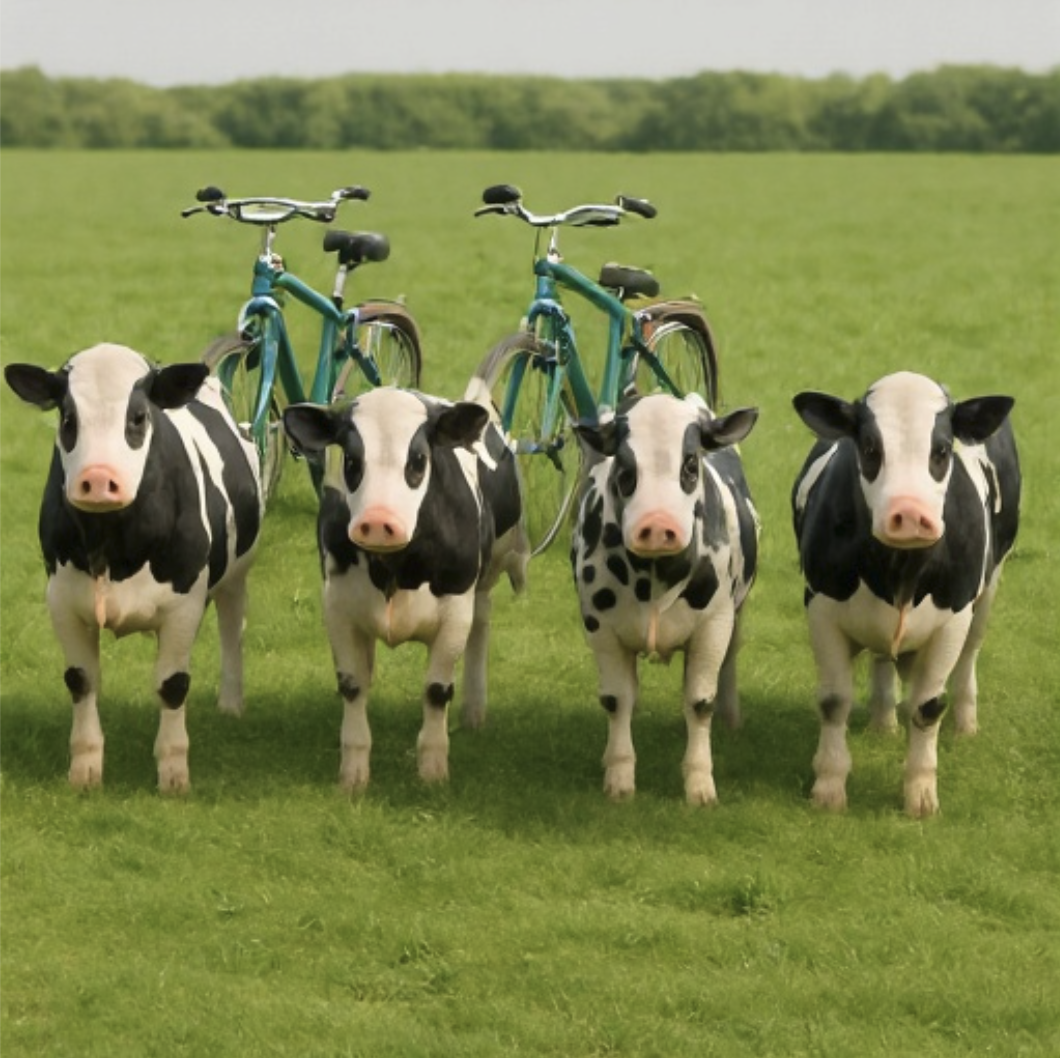}
& \includegraphics[width=0.29\textwidth]{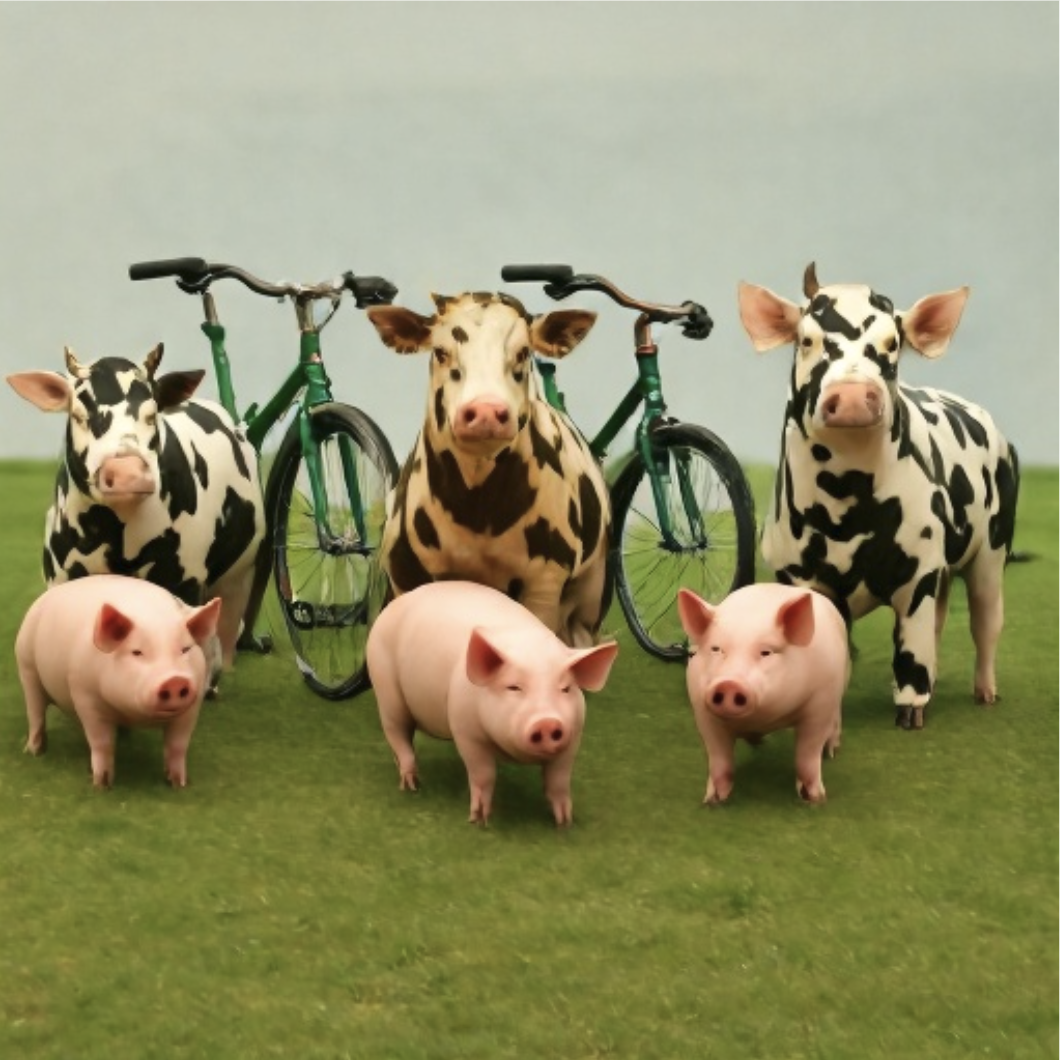}
\\
\end{tabular}
\caption{
Representative failure case at atomicity 8 on BAGEL. 
Prompt: \textbf{``three checkered cows, and two bicycles behind four pigs.'' }
The base UMM and LoRA fine-tuning omit the ``four pigs'' phrase entirely, 
while SST generates pigs but with an incorrect count (three instead of four). 
}
\label{fig:failure}
\end{figure}

We present a representative failure case at high atomicity to complement the quantitative trends reported in Section~\ref{sec:experiments}. 
Across all backbones, generation under dense compositional prompts at high atomicity remains challenging even with our framework. 
Figure~\ref{fig:failure} illustrates this regime through one such case.

%========================================================
\section{Limitations}
\label{app:limitations}

SST inherits its limits from the components it relies on. 
It tunes LoRA parameters only, so its gains are bounded by the generative quality of the backbone. 
Its supervision comes from support masks that contain errors we cannot identify without phrase-level ground truth, and all reported results include these errors. 
Its scaffold modulates a continuous intermediate state, so purely autoregressive image-token UMMs would require a different design.

Two questions remain open. 
Generation at high atomicity stays difficult on all backbones (Appendix~\ref{app:failure}), which we attribute in part to the scarcity of dense prompts in current post-training data. 
Each training pair already includes an image that satisfies its caption, so denser prompts can be constructed and paired the same way. 
We also schedule the scaffold over training progress and do not analyze its contribution across diffusion timesteps.

%========================================================
\section{Broader Impact}
\label{app:impact}
% Positive: reduced annotation cost, more accessible compositional generation
% Negative: compositional generation may aid deepfake / disinformation pipelines
% Mitigation considerations
Our work focuses on improving compositional fidelity in unified multimodal models, a foundational research direction with no immediate deployment path. 
Improvements in text-to-image generation more broadly carry well-documented risks, including the potential for misinformation, deepfakes, and copyright or privacy violations. 
We do not introduce new generative capabilities beyond those of the base UMMs we build upon, and we do not release new model weights; our framework only modifies the training procedure. Standard mitigations such as content provenance, watermarking, and access controls remain applicable to any downstream deployment of the base models.

%%%%%%%%%%%%%%%%%%%%%%%%%%%%%%%%%%%%%%%%%%%%%%%%%%%%%%%%%%%%

%\clearpage
%\newpage
%\input{checklist.tex}

\end{document}